\documentclass[twocol]{ametsocV6.1}
\usepackage{amsmath}    
\usepackage{amssymb}    
\usepackage{bm}         
\usepackage{hyperref}

\newcommand{\diff}{\mathrm{d}}              
\newcommand{\xx}{\bm{x}}                    

\newcommand{\nablaxx}{\nabla_{\bm{x}}}      
\newcommand{\signal}{\boldsymbol{y}}

\newcommand{\noise}{\boldsymbol{n}}
\newcommand{\pdata}{p_\text{data}}
\newcommand{\boldzero}{\mathbf{0}}
\newcommand{\boldi}{\mathbf{I}}

\newcommand{\cskip}{c_\text{skip}}
\newcommand{\cout}{c_\text{out}}
\newcommand{\cin}{c_\text{in}}
\newcommand{\cnoise}{c_\text{noise}}

\usepackage{xcolor}

\newcommand{\hide}[1]{ }

\title{How to use score-based diffusion in earth system science: A satellite nowcasting example}

\authors{Randy J. Chase\aff{a,b}\correspondingauthor{Randy Chase, dopplerchase12@gmail.com}, Katherine Haynes\aff{a}, Lander Ver Hoef\aff{a}, Imme Ebert-Uphoff\aff{a,c},}

\affiliation{
\aff{a}{Cooperative Institute for Research in the Atmosphere, Colorado State University, Fort Collins, CO}\\
\aff{b}{Tomorrow.io, 9 Channel Center St, Boston, MA 02210}\\
\aff{c}{Electrical and Computer Engineering, Colorado State University, Fort Collins, CO}
}

\abstract{
Machine learning (ML) is used for many earth science applications; however, traditional ML methods trained with squared errors often create blurry forecasts. Diffusion models are an emerging generative ML technique with the ability to produce sharper, more realistic images by learning the underlying data distribution. Diffusion models are becoming more prevalent, yet adapting them for earth science applications can be challenging because most articles focus on theoretical aspects of the approach, rather than making the method widely accessible.
This work illustrates score-based diffusion models with a well-known problem in atmospheric science: cloud nowcasting (zero-to-three-hour forecast). After discussing the background and intuition of score-based diffusion models using examples from geostationary satellite infrared imagery, we experiment with three types of diffusion models: a standard score-based diffusion model (Diff); a residual correction diffusion model (CorrDiff); and a latent diffusion model (LDM). Our results show that the diffusion models not only advect existing clouds, but also generate and decay clouds, including convective initiation. A case study qualitatively shows the preservation of high-resolution features longer into the forecast than a conventional U-Net. The best of the three diffusion models tested was the CorrDiff approach, outperforming all other diffusion models, the conventional U-Net, and persistence. The diffusion models also enable out-of-the-box ensemble generation with skillful calibration. By explaining and exploring diffusion models for a common problem and ending with lessons learned from adapting diffusion models for our task, this work provides a starting point for the community to utilize diffusion models for a variety of earth science applications.
}

\begin{document}

\maketitle

%
%
%
\statement

Machine learning is an invaluable tool for earth science applications, but they often result in blurry images, predictions, or forecasts.  Diffusion models are an emerging technique to enable more realistic looking images. This work intuitively explains the diffusion modeling process and explores diffusion models for forecasting satellite imagery. Our results show skillful performance by diffusion models, outperforming traditional machine learning techniques. We discuss lessons learned from applying different diffusion models and trade-offs between performance and computing requirements that are important considerations for deploying these models. The main goal is to provide basic intuition for diffusion modeling and to pass along our hard-earned tips to help accelerate others towards using diffusion models for their own Earth and Environmental Science research tasks.

%


\section{Introduction}

The use of machine learning in meteorology has been growing rapidly \citep[c.f., Figure 1 in ][]{Chase2022}; however, the challenge with neural networks trained with mean squared error losses is that while they are skillful, they often result in blurry predictions \citep{bouallegue2024rise}.
One way to combat the blurry issue has been the use of generative AI methods, such as Generative Adversarial Networks \citep[GANs; ][]{Goodfellow2014,Andrianakos2019,Ravuri2021}. Unfortunately, GANs are notorious for being difficult to train \citep{Saxena2021} and have suboptimal uncertainty quantification \citep[e.g., under-dispersive ensemble forecasts; ][]{Harris2022,Price2022}. An emerging generative AI method to combat blurry output is diffusion modeling \citep{ho2020,Song2021,Karras2022}. Diffusion models use machine learning to iteratively solve a carefully constructed differential equation (inspired by the mathematical equations governing the physical process of heat diffusion, thus the name) to turn random noise into generated text or images that are consistent with the distribution of the data they are trained on. 

Diffusion models have much potential for Earth System science applications; however, one challenge to their widespread use is that they build on complex mathematical concepts and there are few resources available for earth system scientists to easily learn those concepts.
While numerous papers in the Earth System science literature \textit{use} diffusion models, they generally refer readers to computer science literature to \textit{learn} about diffusion models. Currently, the main resources to understand diffusion models are the original computer science papers \citep[e.g.,][]{Song2021,Karras2022}, which take considerable time to digest and to put into practice. 
The goal of this paper is thus to provide an accessible introduction to diffusion models which is tailored to domain scientists in atmospheric science in the hope of making diffusion models more accessible to this community.  
Our approach consists of:
\begin{enumerate}
\item 
    Providing a self-contained, easy-to-understand introduction of the key foundational concepts of diffusion models. Specifically, we focus here on \textit{score-based} diffusion models \citep{Song2021,Karras2022}, which to date are the most commonly used type of diffusion models in atmospheric science.
\item
    Illustrating the use of diffusion models for a well-known dataset and problem in atmospheric science. Namely, we explore their use to nowcast infrared brightness temperatures of geostationary satellites. The end goal of such a nowcasting model is eventually to forecast clouds and precipitation, but here it is used primarily for illustrative purposes.
\item 
    Using the application of nowcasting satellite imagery to illustrate the numerous trade-offs and corresponding design decisions one has to navigate when using diffusion models.  
\end{enumerate}
In the exploration of this relatively new methodology, we discuss the hurdles encountered in implementing score-based diffusion models within the Earth System science community, including some often overlooked steps needed for meteorological applications that are not addressed in the computer science literature on diffusion models. The goal of this added discussion and open-access code is to accelerate the use of score-based diffusion models among domain scientists in the Earth System sciences. 

The application we are utilizing to explore diffusion modeling is cloud and precipitation nowcasting. Clouds and precipitation play a vital role in understanding the radiative budget of Earth and understanding the past, current, and future climate. Furthermore, the exact location and timing of clouds and precipitation have natural applications to forecast weather, solar power generation \citep{Zhang2018} and laser communications \citep{Craddock2024}. The main methods for forecasting clouds and precipitation are Lagrangian advection schemes \citep[e.g.,][]{Bowler2006,Pulkkinen2019}, traditional numerical weather prediction \citep[e.g.,][]{Griffin2017,Griffin2024}, and optical flow \citep[e.g.,][]{Shakya2019}. Recently, machine learning has been rapidly employed for forecasting clouds and precipitation using 
traditional deep learning techniques \citep[e.g.,][]{Shi2015,Berthomier2020,Caseri2022,Kellerhals2022}, 
generative adverserial models \citep[e.g.,][]{Andrychowicz2023}, 
and diffusion modeling \citep{Hatanaka2023,Zhang2023,Dai2024,Nai2024,Wang2024,Yu2024}.
In addition to becoming prevalent in nowcasting, diffusion models are also being used more generally in data-driven weather forecasting \citep{Couairon2024,Manshausen2024,Pathak2024,Price2025}. 

Here, we are tasking the model with forecasting geostationary satellite infrared (IR) imagery, focusing on nowcasting (ten minutes out to three hours) 10.3 $\mu m$ brightness temperatures.
In our exploration of score-based diffusion models we try three types that all follow the main design approach from \citet{Karras2022}, also known as Elucidating the Design space of diffusion-based generative Models (EDM). The three specific types are
\begin{enumerate}
\item 
   A plain diffusion model \citep[Diff; ][]{Karras2022},
\item
   A residual correction diffusion model \citep[CorrDiff; ][]{Mardani2025,Pathak2024}, 
\item
   A latent diffusion model \citep[LDM; ][]{Rombach2022,Leinonen2023}.
\end{enumerate}

This paper is organized as follows. 
Section \ref{sec:intro_to_diff} contains an intuitive overview of score-based diffusion, including a brief discussion of the mathematical foundations of diffusion models. 
Section \ref{sec:data_and_methods} provides background on the geostationary satellite dataset we use, the steps to prepare it for using score-based diffusion, how it is adapted for the satellite forecasting task, what specific types of machine learning models are trained, and some exploration of pre-trained autoencoders. 
Section \ref{sec:results} shows the results of several score-based diffusion models trained to nowcast satellite imagery and discusses their performance compared to a couple of baseline forecasts, persistence and a mean squared error trained U-Net. 
Section \ref{sec:discussion} discusses lessons learned, limitations, and improvements that could be made.
Lastly, Section \ref{sec:conclusions} summarizes the paper.

\section{Intuition for Score-Based Diffusion Models}
\label{sec:intro_to_diff}

\subsection{Preface}
All diffusion models discussed in this paper follow a score-based diffusion framework \citep[EDM; ][]{Song2021,Karras2022}. Before we dive into the details of the EDM framework we acknowledge another common framework for diffusion models, named Denoising Diffusion Probabilistic Models \citep[DDPM; ][]{ho2020}. We initially explored DDPMs for this project given their wide use in the literature \citep[e.g.,][]{Asperti2024,Chen2023,Gao2023,Srivastava2024,Zhong2024}, but realized the computational demand would not be viable given our hardware and the temporal demand of an operational nowcasting product. Instead, after seeing the results from \citet{Li2024}, \citet{Mardani2025}, \citet{Pathak2024}, and \citet{Price2025}, we decided to pursue score-based diffusion models which we found easier to train and evaluate. The EDM framework can be interpreted as a generalization of the DDPM approach. For more information on this topic, see \citet{ho2020} for DDPMs, see \citet{Song2021} and \citet{Karras2022} for the connection between the DDPM and EDM framework, and see references therein and several Earth and Environmental Science papers using DDPMs \citep{Asperti2024,Chen2023,Gao2023,Srivastava2024,Zhong2024}. Similarly, readers may have heard of flow matching or continuous normalizing flow \citep{Lipman2023,Tong2024} which can be thought of as a more abstract version of diffusion models. Both define a way that maps noise to data but differ in how the network is trained and the underlying choice of formulation. For a technical description of flow matching with an application to weather forecasting see \citet{Stock2025}.

\subsection{Basic Idea of Diffusion Models}
\label{sec:diff_basic_idea}

Generally, diffusion models can be thought of as tools to learn, and later sample from, data distributions. A diffusion model is trained on a large set of representative samples of the distribution, typically images. For example, if a diffusion model is trained on an extensive set of sample images from a specific satellite channel, the diffusion model learns to approximate this distribution. Once trained, the diffusion model can provide new samples from the distribution, i.e., it generates new (approximate) satellite imagery for that channel. This type of random but realistic image generation is called {\it unconditional} image generation. Although such ability is useful for many applications, we are more interested here in the ability to generate specific imagery, known as {\it conditional} image generation. Conditional image generation would allow, for example, to use information from a different channel of imagery or to forecast satellite imagery 10 minutes from now by conditioning on relevant imagery. 
In Sections \ref{sec:intro_to_diff}\ref{sec:diff_basic_idea} to \ref{sec:intro_to_diff}\ref{sec:image_generation} we focus entirely on diffusion models for unconditional imagery, while Section \ref{sec:intro_to_diff}\ref{sec:conditional_diffusion} discusses how to add the conditioning. 

To perform image generation, diffusion models utilize a mathematically well-described diffusion process for both the training of the underlying neural network and generating new images. Namely, to learn the distribution from a given data set of images (training set), training images are {\it partially destroyed} by applying a diffusion process to the image (i.e., adding noise). This is called the ``forward process''.
The diffusion model is then trained to {\it restore} the structure of the images from their degenerated versions. 
This is called the ``backwards process'', and it teaches the model to generate imagery of the same structure as the images in the training data set. In other words, the model learns to approximate the distribution represented by the training data set. 
Diffusion models are thus based on developing a model to reverse the diffusion equation, an approach developed mathematically over 40 years ago by \citet{anderson1982reverse}. However, certain components of this approach are hard to implement in practice, and thus the full potential of this approach was made possible only once those components could be implemented through neural networks, as proposed by  \citet{sohl2015deep}.

\subsection{Score-based Diffusion - Mathematical Foundation}
\label{sec:diff_math}

Score-based diffusion is, in essence, a distribution-to-distribution transformation. 
We denote as $\mathcal{X}$ the distribution from which we want to generate new data (typically images, and in our case GOES IR images over time).  $\mathcal{X}$ is the abstract distribution from which the training data was drawn, consisting of elements $x$.  The training data is denoted as $\hat{\mathcal{X}}$ which we use as an approximation of $\mathcal{X}$. The probability density function for $\mathcal{X}$, $p(x)$, describes how representative or uncommon a particular $x$ is within the scope of $\mathcal{X}$; to make this concrete, for our example, in theory \emph{any} combination of pixel values of the appropriate shape could be observed from GOES, but some combinations are far more likely than others, and that \emph{likelihood} is what $p(x)$ represents. 

\begin{figure*}[t]
 \begin{center}
 \includegraphics[width=6in]{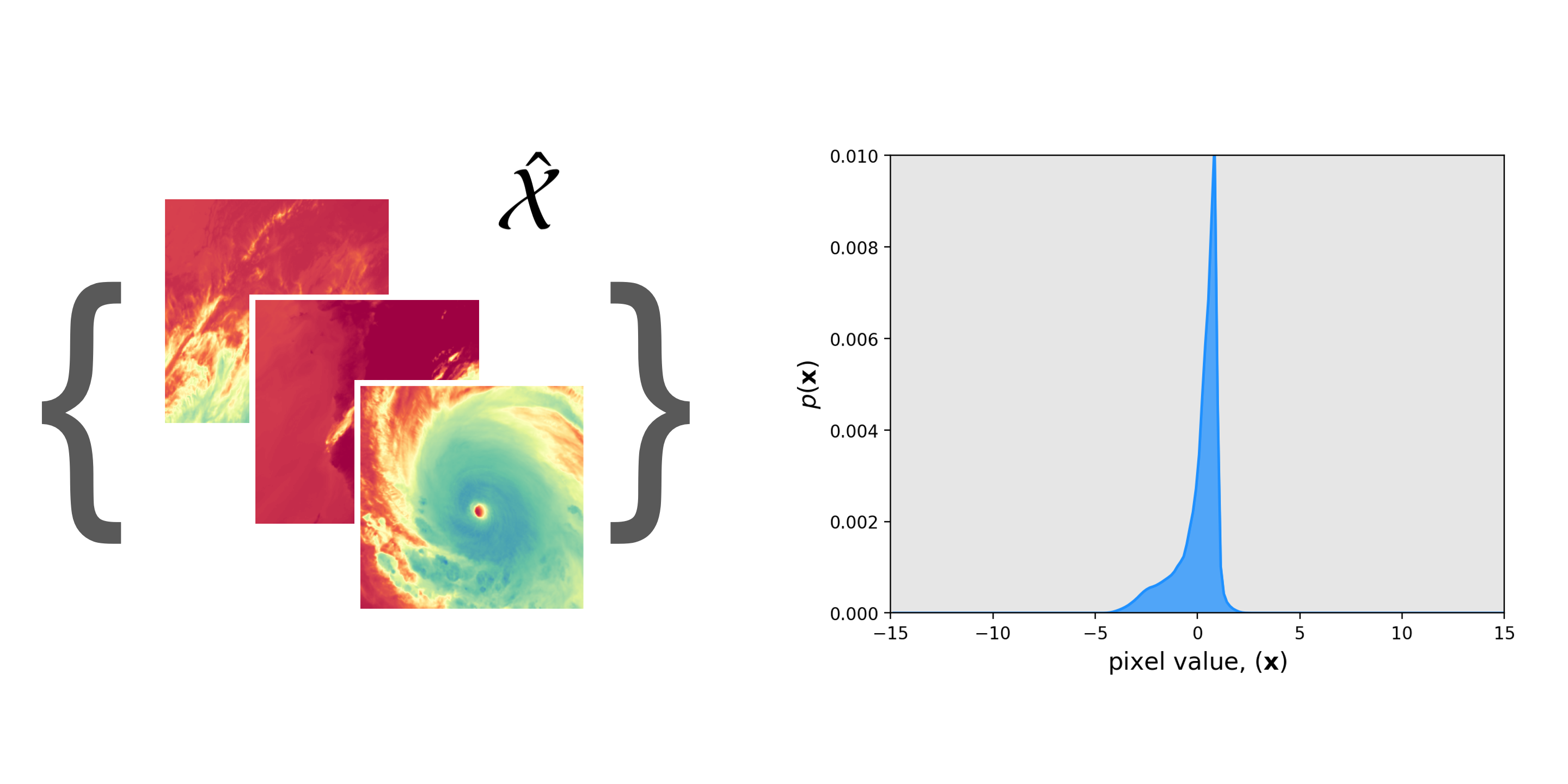}
 \caption{Graphic illustration of what the distribution of training data looks like, both in its 2d shape as images (left) and a 1d representation for conceptualization (right, the histogram representing the pixel-value distribution).}\label{fig:Fig_1}
 \end{center}
\end{figure*}

Fig.\ \ref{fig:Fig_1} shows on the left sample images from $\hat{\mathcal{X}}$ and on the right a simplified representation of an estimation of $p(x)$ as a histogram. While $\hat{\mathcal{X}}$ is a high-dimensional distribution with complex spatial relationships, for simplicity of visualization the histogram shown is 1-dimensional and represents the distribution of the values for a single pixel (i.e., for a ``1x1'' image). 

\begin{figure*}[t]
 \begin{center}
 \includegraphics[width=6in]{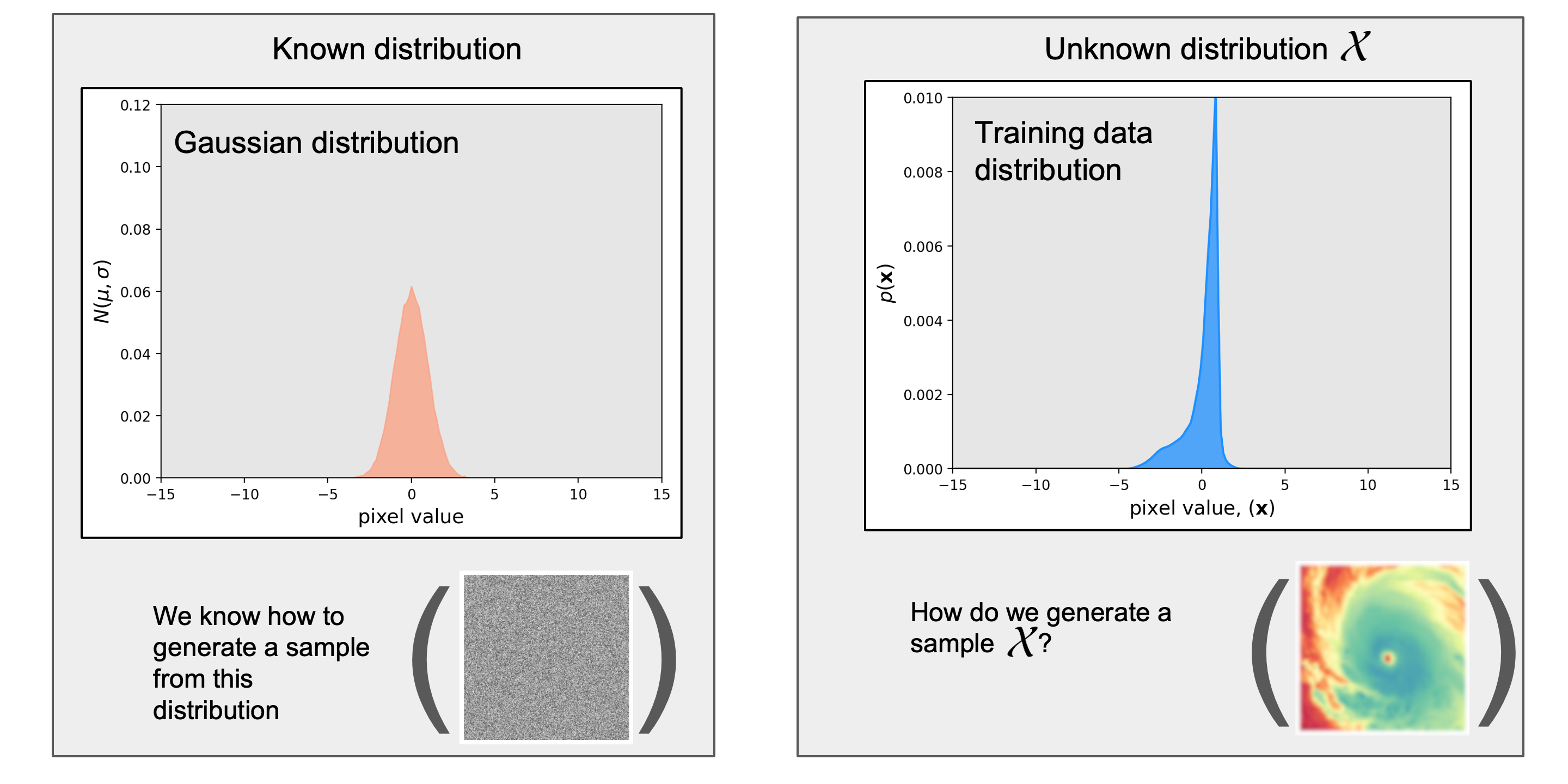}\\
 \caption{Schematic describing main task for diffusion models:  seeking a way to sample from an unknown distribution, ${\mathcal{X}}$, by sampling from a known distribution, such as a Gaussian distribution, $\mathcal{N}(\mu,\sigma)$.}
 \label{fig:Fig_2}
 \end{center}
\end{figure*}

The simplified plot of $p(x)$ in Fig.\ \ref{fig:Fig_1} notwithstanding, we have neither an explicit formulation for $p(x)$ nor knowledge of how to draw a new sample image from $\mathcal{X}$. However, we do know how to draw samples from many well-known distributions such as a Gaussian distribution $\mathcal{N}(\mu, \sigma)$ (bell-shaped curve), shown as the ``known distribution'' in Fig.\ \ref{fig:Fig_2}. Moreover, because the Gaussian distribution is closely related to ``white noise'', we can describe a process to iteratively transform images from $\hat{\mathcal{X}}$ into samples from a particular multivariate Gaussian by adding sufficient quantities of white noise; that is, we know how to map from the ``unknown distribution'' to the known distribution in Fig.\ \ref{fig:Fig_2}.
The goal of diffusion is to learn the inverse of that mapping: to learn how to map a sample from the known distribution (which is easy to generate) to a corresponding sample in the unknown distribution $\mathcal{X}$.

\begin{figure*}[t]
 \begin{center}
 \includegraphics[width=6in]{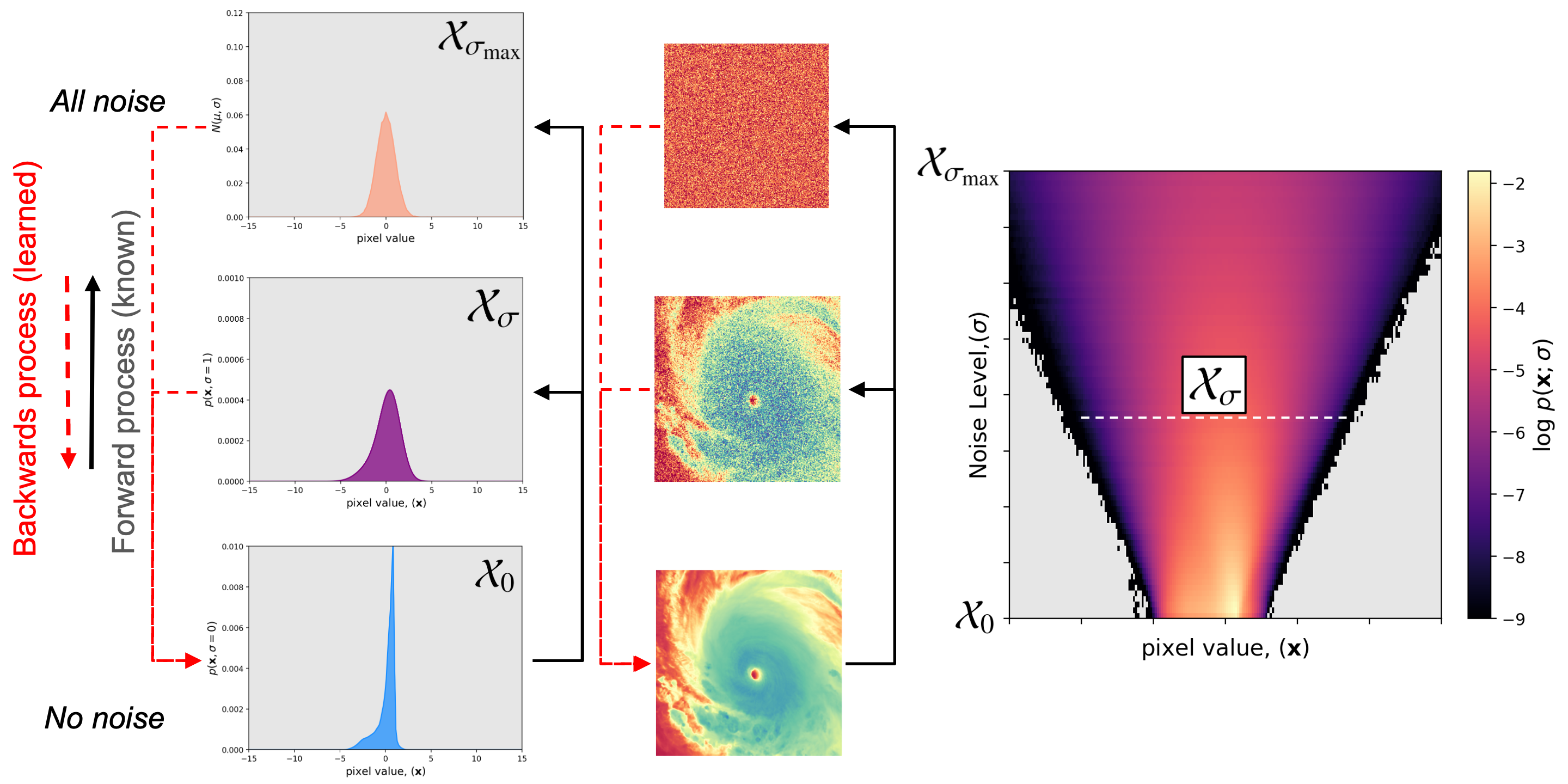}\\
 \caption{A visual diagram illustrating conceptually the forward and backward process for a diffusion model. The left column shows the simplified 1-d representations of distributions. The center column shows single image examples. The right column shows all 1-d representations of the data, which is known as $x-\sigma$ diagram in the computer science literature. Note that we switched the axes of the $x-\sigma$ diagram - in comparison to the literature - to emphasize the relationship to the histograms in the left column. Namely,  each histogram on the left represents one $\sigma$ value and corresponds to one horizontal line in the $x-\sigma$ diagram on the right.}
 \label{fig:Fig_3}
 \end{center}
\end{figure*}

In Fig.\ \ref{fig:Fig_3}, we illustrate how we can create these mappings by creating intermediate spaces with increasing amounts of noise contamination between the known and unknown distributions, which we now introduce notation for: what we have been calling $\mathcal{X}$ is now $\mathcal{X}_0$ (where the subscript indicates noise level 0), for each noise level $\sigma$ we have $\mathcal{X}_\sigma$ up to some maximum $\sigma = \sigma_{\max}$.
The \emph{forward process} starts with an intact image from $\hat{\mathcal{X}}$ (bottom of Fig.\ \ref{fig:Fig_3}) and adds Gaussian noise repeatedly until only noise is left (top of Fig.\ \ref{fig:Fig_3}, where $\mathcal{X}_{\sigma_{\max}} \approx \mathcal{N}(\mu, \sigma_{\max})$). This is straight forward and efficient (as we can combine steps to jump directly from a sample in $\mathcal{X}_0$ to a corresponding sample in $\mathcal{X}_\sigma$ for any $\sigma$ in one step), and we make use of this to expand our training set $\hat{\mathcal{X}}$ into a continuum of increasingly noisy distributions $\hat{\mathcal{X}_\sigma}$ (which in turn approximate the abstract distributions $\mathcal{X}_\sigma$). The 1D histograms for one of these spaces are shown in the left column of Fig.\ \ref{fig:Fig_3}, and the complete set of 1D histograms are plotted as the horizontal slices of the rightmost plot in Fig.\ \ref{fig:Fig_3}. This ``$x$-$\sigma$'' diagram is an illustration of the space we have created to interpolate between the known and unknown distributions.

The \emph{backwards process} is the hard part and we need machine learning to do this. The goal is to turn an image of random noise (top of Fig.\ \ref{fig:Fig_3}) into an image that would be within the distribution of ${\mathcal{X}}$ (bottom of Fig.\ \ref{fig:Fig_3}), i.e., a GOES IR image in our case. The trick of the backward process is to iteratively remove the noise in the image while guiding each pixel's value so that the modified image is representative of ${\mathcal{X}}$ (GOES IR image). This is an iterative process that is performed in many steps, generating a sequence of images that removes a small amount of noise at each step, which corresponds to a trajectory through the $x$-$\sigma$ space (rightmost figure in Fig.\ \ref{fig:Fig_3}) that steps slowly down through the $\mathcal{X}_\sigma$ layers from the top (known distribution) to the bottom (unknown distribution $\mathcal{X}_0$). The left two columns of Fig.\ \ref{fig:Fig_3} show this process collapsed into two steps for simplicity of visualization. In each iteration, we want to move the sample $\xx$ (new image that we are generating) in a direction where it will become more likely under $\mathcal{X}$ -- that is, we want to maximize $p(\xx)$, and in particular $p(\xx, \sigma)$, which is the same idea of a likelihood, just specialized to a particular noise level.
To know what direction to move in to maximize $p(\xx, \sigma)$, we use the \emph{score function}, which is defined to be $\nablaxx \log p(\xx, \sigma)$. Putting this all together, to define how to update $x$ through these iterations (i.e., as we move along this trajectory), we utilize an ordinary differential equation (ODE):
\begin{equation}
  \label{eq:prob_ode}
  \diff \xx = - \sigma \underbrace{\nablaxx \log p\big( \xx; \sigma}_{\text{score function}} \big) \,\diff\sigma.
\end{equation}

\begin{figure*}[t]
 \begin{center}
 \includegraphics[width=4in]{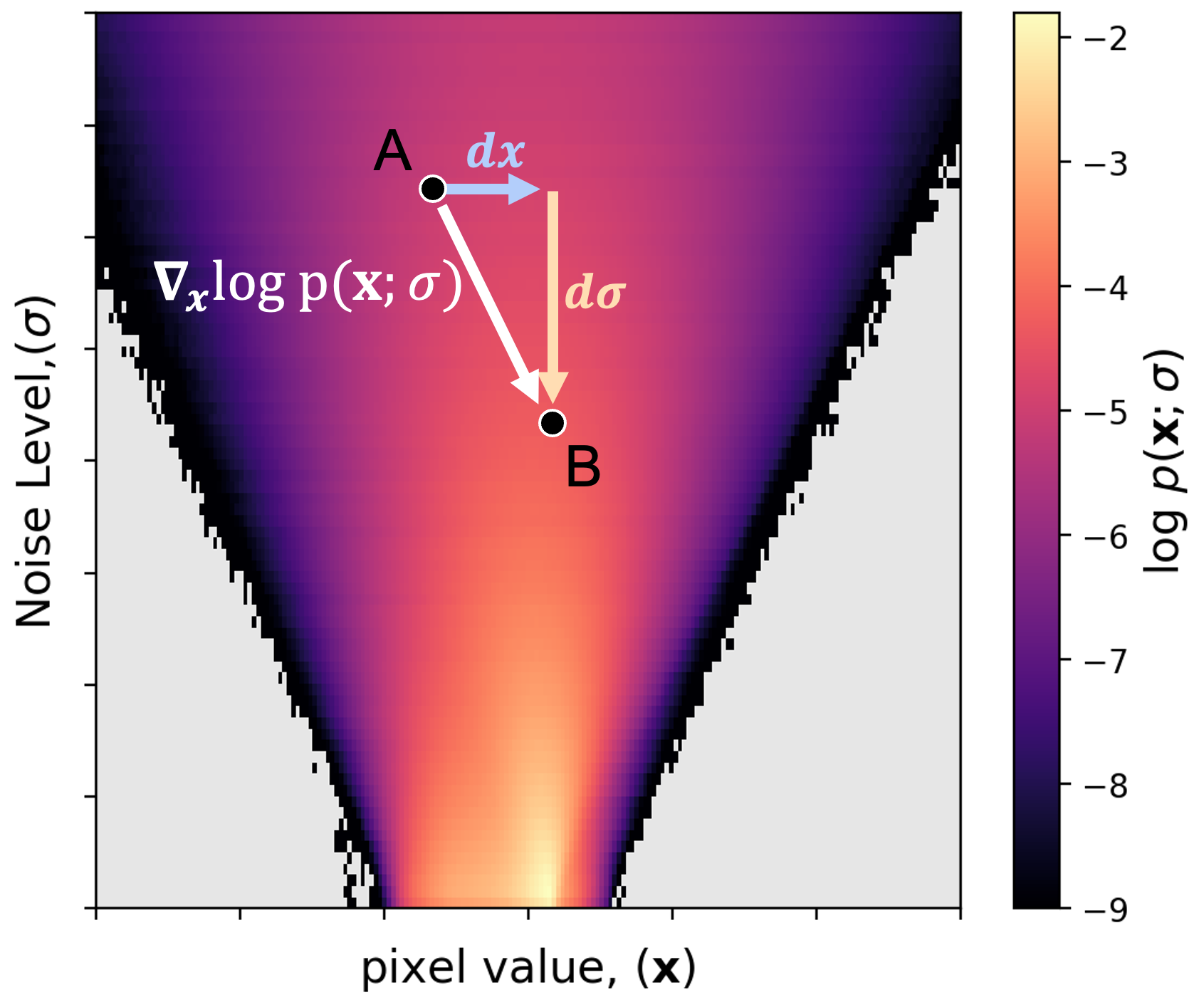}\\
 \caption{$x-\sigma$ diagram illustrating one iteration of the backward process of a diffusion model, namely the transition from one image, A, to the next image, B, that is closer to the distribution of $\mathcal{X}$, using the gradient term, $\nablaxx \log p(\xx, \sigma)$.}
 \label{fig:Fig_4}
 \end{center}
\end{figure*}

Fig.\ \ref{fig:Fig_4} illustrates one step of the process, going from Image A to Image B, where we move from higher noise and lower likelihood (image A) down the gradient to less noise and higher likelihood under ${\mathcal{X}}$ (image B), thus closer to a GOES IR image in our case. We note that elsewhere in the diffusion literature, the reader will likely see the equivalent to Equation \ref{eq:prob_ode} phrased in terms of the variable $t$ and $\sigma$ as a function ($\sigma(t)$) -- this is the more general form, and allows for the noise level to vary non-linearly with the variable being discretized to solve the ODE. For the full mathematical details, we refer interested readers to the foundational paper on score-based diffusion, \citet{Karras2022}.


\subsection{Training the Neural Network}

\citet{Karras2022} proved (in their Appendix B.3) that if one has access to an \textit{ideal denoiser} (i.e., an algorithm that can return an exact copy of a member of the training set given any noised version of that image), then the score function can be computed exactly as
\begin{equation}
\label{eqn:equivalence}
\nablaxx \log p \big( \xx; \sigma) = \frac{D(\xx;\sigma) - \xx}{\sigma^2}
\end{equation}
where the left hand side is the score function and $D(\xx; \sigma)$ is the ideal denoiser. In practice, we do not generally have access to an ideal denoiser, but denoising is a task that neural networks are already known to be well-suited for, so we can leverage existing techniques to approximate this ideal denoiser.
In order to get a \textit{good} denoiser we train a neural network to minimize the loss from denoising an image, which is typically done with the common $L^2$ norm (i.e., Mean Squared Error): 

\begin{equation}
\label{eqn:expectation}
\mathbb{E}_{\signal \sim \pdata} \mathbb{E}_{\noise \sim \mathcal{N}(\boldzero, \sigma^2 \boldi)} \lVert D_\theta(\signal + \noise; \sigma) - \signal \rVert^2_2 .
\end{equation}
Here $\signal$, are the original non-noised images, $D_\theta$ is the neural network denoiser, and  
$\mathcal{N}(\boldzero, \sigma^2 \boldi)$ is a multi-dimensional normal distribution with covariance matrix $(\sigma^2 \boldi)$, where the dimension of the identity matrix, $\boldi$, is the number of pixels in each image. Thus, $\noise \sim \mathcal{N}(\boldzero, \sigma^2 \boldi)$ yields a matrix (image), $\noise$, with independent Gaussian noise (of mean $0$ and variance $\sigma^2$) in each pixel. Furthermore, the expectation, $\mathbb{E}$, symbols in Eq.\ (\ref{eqn:expectation}) indicate that the neural network is trained to minimize the {\it average} taken across all images in the data set and across all noise in the multi-dimensional normal distribution. In practice the loss is also scaled by the current value of $\frac{1}{\sigma}$. The intuition here is that we want to encourage the neural network to perform better at low noise levels (small $\sigma$) because it should be closer to the true solution, so we weight the loss more heavily in that region. 

\begin{figure*}[t]
 \begin{center}
 \includegraphics[width=6in]{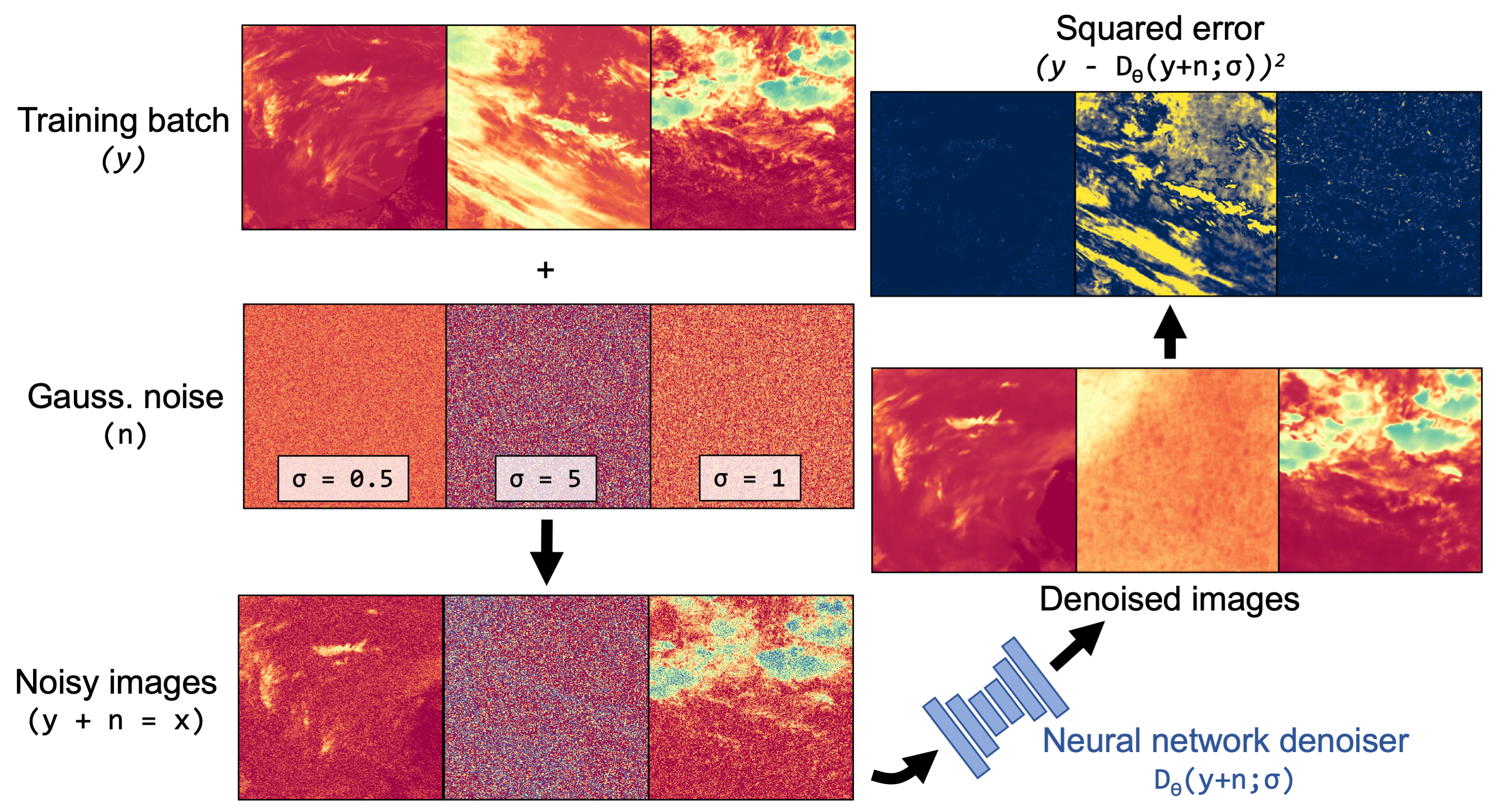}\\
 \caption{A schematic describing one batch of data through the training process of an unconditional diffusion model.}
 \label{fig:Training_Schematic_Unconditional}
 \end{center}
\end{figure*}

Figure \ref{fig:Training_Schematic_Unconditional} is a visual example of a training step for the denoiser. First, we select a batch of satellite images ($\signal$). Then, we draw a random noise level for each image in the batch, $\sigma$, from a distribution defined by the user and use that to generate random Gaussian noise images, $\noise$. The noise is then added directly to the images, yielding ($\signal + \noise$), and  given to the neural network ($D_{\theta}(\signal + \noise; \sigma)$) to be denoised. The choice of neural network architecture is arbitrary, as there have been examples of many different architectures (e.g., U-Nets, vision transformers, graph neural networks) that have been used as successful denoisers. The MSE between the output of the neural network and the original images is then computed and the loss is backwards propagated through the network. This training process is repeated until the loss converges. 

Take note of the double expectation in Equation \ref{eqn:expectation} that implies that good sampling of the images, $\signal$, the noise levels, $\sigma$, and of sampled noise images, $\noise$, are required for training. Thus, in order to get a good estimate of the gradient in your learning process, diffusion will likely need a bigger batch than one would expect compared to using a plain U-Net \citep{Ronneberger2015}, because there needs to be a diverse sampling of examples and noise levels.

The way we have described the training of a denoiser would provide a network sufficient for diffusion, but \citet{Karras2022} developed some clever scalings to make the training task simpler for the neural network. We briefly discuss these scalings here to make it easier to follow the corresponding functions in the code base.
\citet{Karras2022} actually chose a neural network, $F_\theta(\mbox{input}; \mbox{noise level})$, that relates to the denoiser as follows: 
\begin{equation}
D_\theta(\xx; \sigma) = \cskip(\sigma) ~\xx \\ + \cout(\sigma) \underbrace{~F_\theta \big( \cin(\sigma) ~\xx; ~\cnoise(\sigma) \big)}_{\mbox{neural network}} \text{,}
\label{eq:preconditioning}
\end{equation}
where $\cskip(\sigma)$, $\cout(\sigma)$, $\cin(\sigma)$, and $\cnoise(\sigma)$ are all scalar functions of $\sigma$ and transform the data such that training and generation are numerically better behaved and thus can be learned more easily by a neural network. The first function, $\cskip(\sigma)$, is not to be confused with traditional skip connections in residual neural networks or U-Nets. Rather, $\cskip(\sigma)$ is a scalar function that provides a scaled version of $\xx$ such that the errors from $F_\theta$ are not amplified. The scalar functions $\cin(\sigma)$ and $\cout(\sigma)$ ensure that the input and output of the neural network, $F_\theta$, have unit variance. Lastly, $\cnoise(\sigma)$ is a scalar function that maps the original noise value so that it can be used better as a condition for the neural network. The derivation of these functions can be found in Appendix B.6 in \citet{Karras2022}. The functions are as follows, where $\sigma_{data}$ denotes the scalar variance of the training data, calculated across all pixels: 
\begin{equation}
\cskip(\sigma) = \sigma_{data}^2/(\sigma^2 + \sigma_{data}^2) 
\label{eq:c_skip}
\end{equation}
\begin{equation}
\cout(\sigma) = \sigma \cdot \sigma_{data} / 
\sqrt{\sigma_{data}^2 + \sigma^2} 
\end{equation}
\begin{equation}
\cin(\sigma) = 1/ \sqrt{\sigma_{data}^2 + \sigma^2} 
\end{equation}
\begin{equation}
\cnoise(\sigma) = \frac{1}{4}\ln(\sigma)
\end{equation}
We found that these functions derived in \citet{Karras2022} worked well for our satellite nowcasting task and suspect that it could be a good starting points also for many Earth and Environmental Science tasks.

\subsection{Generating Images with the Trained Network}
\label{sec:image_generation}

The goal of a diffusion model for unconditional image generation is to generate new images that belong to the training dataset starting from pure Gaussian noise. Thus, to start generating new images we first sample a pure noise image that has a large $\sigma = \sigma_{\max}$ value (e.g., 80). In order to translate this pure noise image into an image from the training data we leverage the trained network,  $F_\theta$, and subsequently,   $D_\theta$, by substituting the equivalence in Equation \ref{eqn:equivalence} into Equation \ref{eq:prob_ode}
\begin{equation}
\label{eqn:sub1}
\diff \xx = -\sigma\frac{D_\theta(\xx;\sigma) - \xx}{\sigma^2} ~\diff\sigma.
\end{equation}
Canceling the $\sigma$ yields
\begin{equation}
\label{eqn:sub2}
\diff \xx = - \frac{D_\theta(\xx;\sigma) - \xx}{\sigma} ~\diff\sigma.
\end{equation}
Notice how Equation \ref{eqn:sub2} effectively states that in order to move the pure noisy image from higher noise to lower noise we can use the slope between the denoised image out of the neural network and the original noisy image. Equation \ref{eqn:sub2} is written in its continuous form and can be discretized into 

\begin{equation}
\label{eqn:discrete}
(\xx_i - \xx_{i-1}) = - \frac{D_\theta(\xx_i;\sigma_i) - \xx_i}{\sigma_i} (\sigma_i - \sigma_{i-1})
\end{equation}
which results in 

\begin{equation}
\label{eqn:discrete_final}
\xx_{i-1} =  \xx_i + \big(\frac{D_\theta(\xx_i;\sigma_i) - \xx_i}{\sigma_i}\big)(\sigma_i - \sigma_{i-1})
\end{equation}
The subscript, $i$ denotes the current noisy step, and $(i-1)$ denotes moving backwards towards the original data distribution, i.e., towards the goal. 

\begin{figure*}[t]
 \begin{center}
 \includegraphics[width=6in]{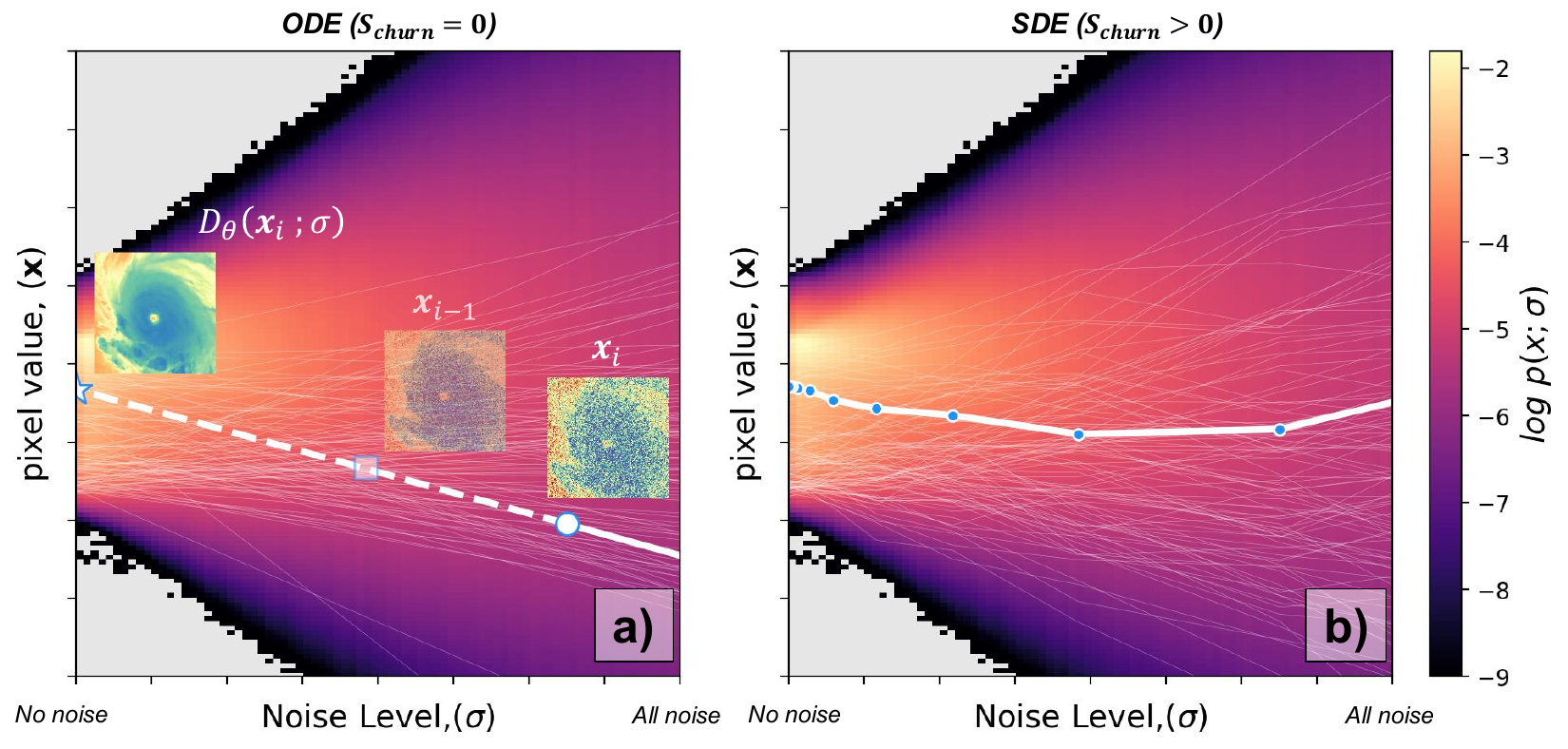}\\
 \caption{One step in image generation using score-based diffusion. (a) One step of denoising an image ($\xx_i$). The background shading is the same as Figure \ref{fig:Fig_4} but the image has been rotated to match other literature examples. The circle marker is the location of one pixel within $\xx_i$, the star is the denoised value of that same pixel. The transparent image is the result of one step of removing noise. The other white lines are 256 random pixels of $\xx$ and their journey through denoising them. (b) same as in (a) but the trajectories now follow a stochastic trajectory.}
 \label{fig:image_generation_x_sigma}
 \end{center}
\end{figure*}

Figure \ref{fig:image_generation_x_sigma}a shows an example of the image generation starting at $\sigma = 3$ (circle marker). The first step is to take the noisy image ($\xx_i$) and plug it into $D_\theta(\xx_i;\sigma_i)$, which gives us the denoised image located at the star. We can now calculate the slope between the circle and star $\left(\frac{D_\theta(\xx_i; \sigma_i) - \xx_i}{\sigma_i}\right)$. Plugging into Equation \ref{eqn:discrete_final} we can take a step of removing noise along this slope and end up at $\xx_{i-1}$ (transparent square and image in Figure 3a). This is repeated until we reach the left hand axis. This is an oversimplified example because the actual markers and lines in Fig.\ \ref{fig:image_generation_x_sigma} represent changes in individual pixels in the image, $\xx$. Thus, along a single trajectory in an abstract $256 \cdot 256$-dimensional space.

Notice that despite updating their slope at each $\sigma_i$ value, the lines in Fig.\ \ref{fig:image_generation_x_sigma}a are relatively straight trajectories (faint lines in background). Moreover, because the distribution $p(x; \sigma)$ is fixed, these trajectories are completely determined by their origin on the right $y$-axis (i.e., their initial random seed). To add more randomness and variation to the generation process, random walks are incorporated by injecting a user defined (i.e., hyperparameter) amount of noise during the denoising procedure. Effectively, think of the same generation procedure as in Fig.\ \ref{fig:image_generation_x_sigma}a, but after we generate $\xx_{i-1}$ we actually add a bit more random noise in (i.e., small step in a random direction towards the right y-axis). This leads the image generation to deviate from the straight ODE paths, and enhances the ensemble image diversity across samples. An example of adding in the stochastic noise during the generation procedure is in Fig.\ \ref{fig:image_generation_x_sigma}b. Trajectories now curve on their journey from the right y-axis to the left y-axis. In the \citet{Karras2022} framework and the associated code, the scale of the noise added back in is named $S_{churn}$ and the amount of noise is named $S_{noise}$.

\subsection{Adding a Condition}
\label{sec:conditional_diffusion}

\begin{figure*}[t]
 \begin{center}
 \includegraphics[width=6in]{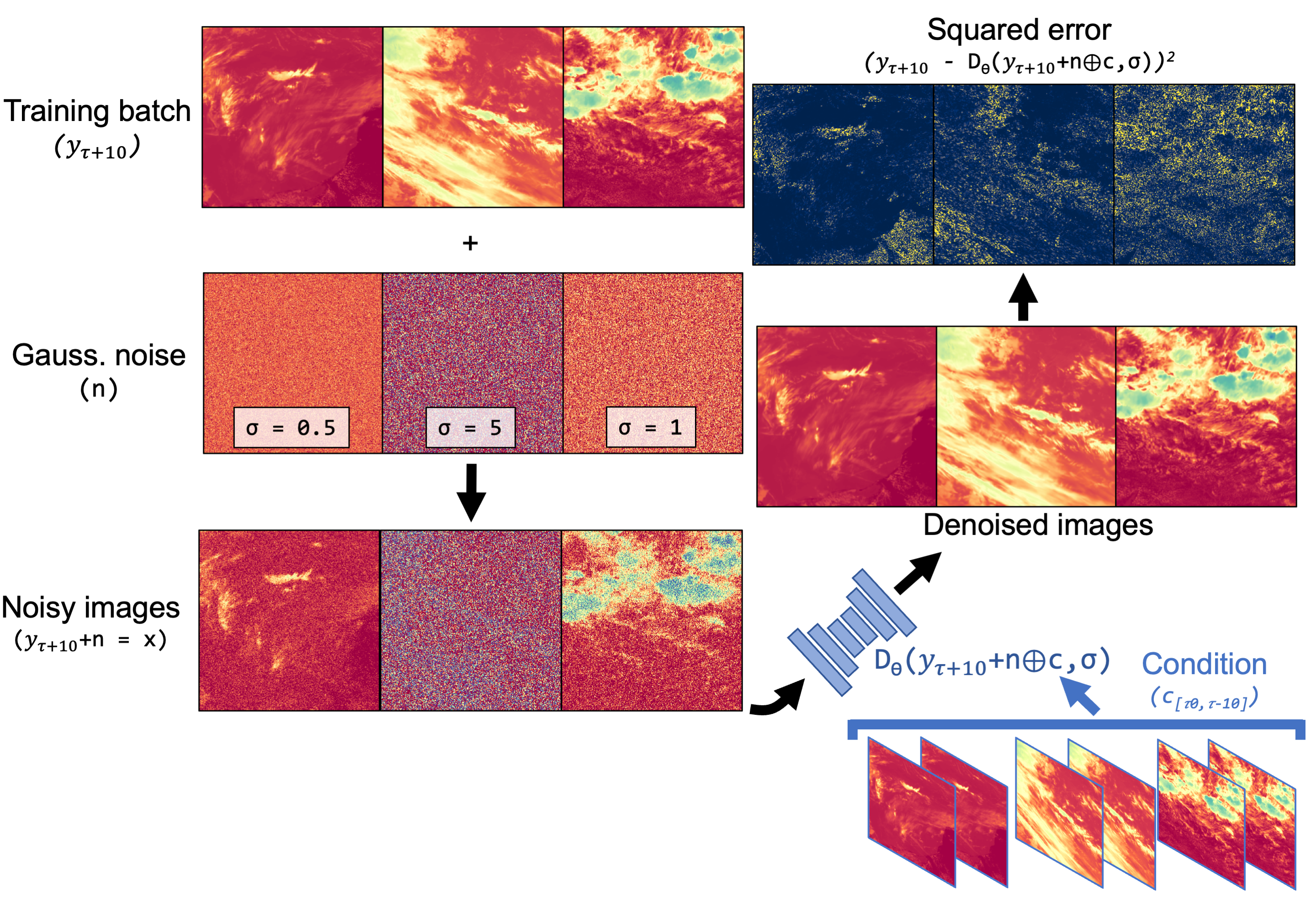}\\
 \caption{A schematic describing one batch of data through the training process of a conditional diffusion model.}
 \label{fig:Training_Schematic_Conditional}
 \end{center}
\end{figure*}

The description thus far has described \textit{unconditional} image generation, that is, generating a random GOES IR image. Many tasks in the Earth and Atmospheric Sciences are much more aligned with \textit{conditional} image generation. For our paper the task is to generate a specific GOES IR image, the image 10 minutes from now. Most papers on diffusion in the computer science literature tend to build diffusion models for text-to-image tasks \citep[Dall-e , ImageGen; ][ respectively]{Ramesh2021,Saharia2022}. For these text-to-image tasks, incorporating the condition (i.e., the text prompt) is usually done by feeding it into the neural network as a text embedding (i.e., some sort of encoded vector of text). Typically the text embedding relies on known methods to translate the text prompt into tokens and a matrix of data. 

\begin{figure*}[t]
 \begin{center}
 \includegraphics[width=6in]{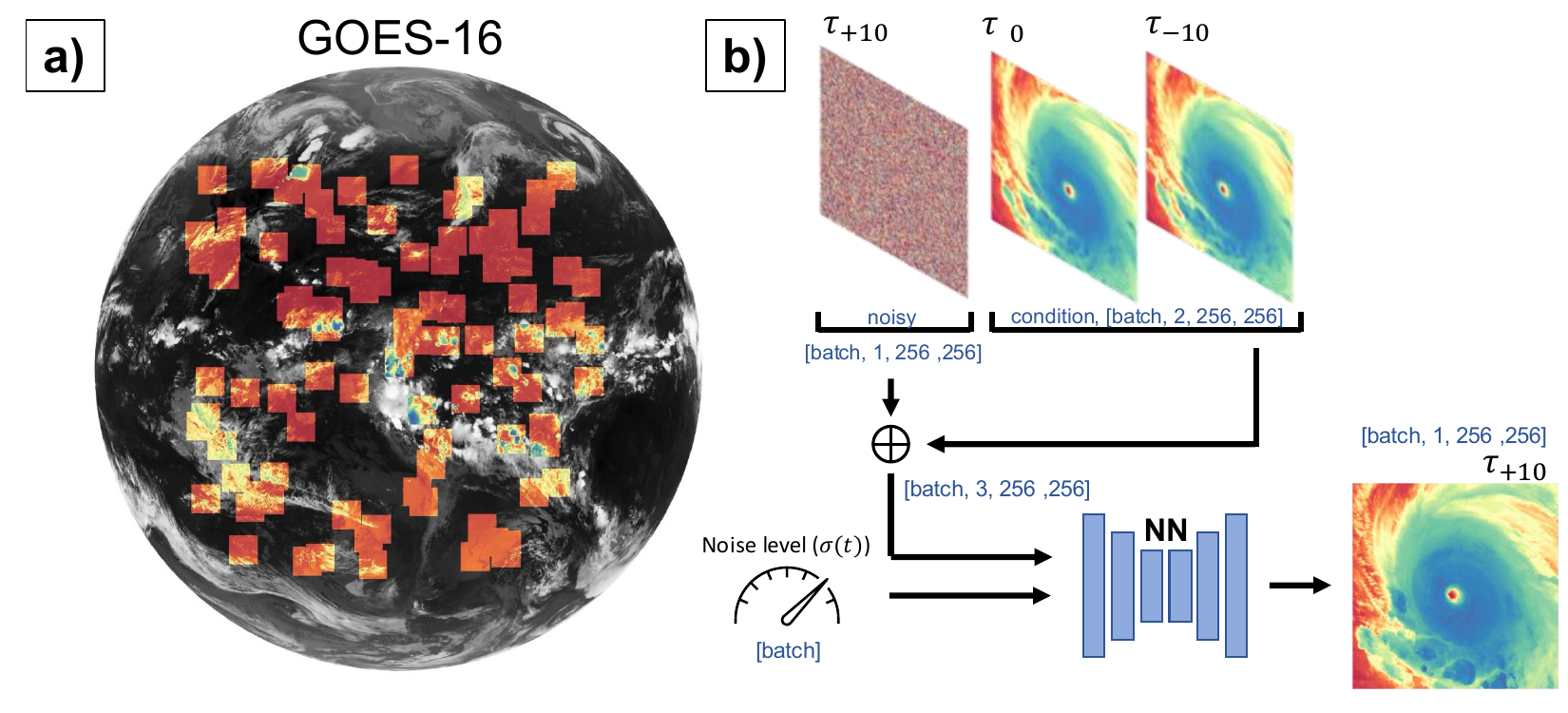}\\
 \caption{Visual example of the GOES-16 patching and the general flow of data through our neural network. More specifically, this is an example GOES-16 IR image from Channel 13 (shown as black and white background shading) with random 256 by 256 pixel patches selected for the training and test dataset (shown in color). (b) The shapes of data going into the neural network used for the diffusion model. $\tau$ represents the time of the forecast, where $\tau_{+10}$ at the top of the image is a noisy version of the 10 min forecast and $\tau_0$ is the current time and $\tau_{-10}$ is the previous 10 min image. The circle with cross is concatenation of the two sets of images. The rectangles with the NN (neural network) label represent the underlying U-Net for the denoiser.}
 \label{fig:GOES_data_patches}
 \end{center}
\end{figure*}

For our task, image-to-image translation, it was not initially clear how to include feature images as a condition. Through some experimentation we found that including our condition (the past 2 GOES IR images) as additional channels to the input (i.e., concatenate the condition to the noisy dimension) to the denoiser U-Net worked well (Fig.\ \ref{fig:Training_Schematic_Conditional} and Fig.\ \ref{fig:GOES_data_patches}b).
Comparing the results of the unconditional model in Fig.\ \ref{fig:Training_Schematic_Unconditional} to the results of the conditional model in Fig.\ \ref{fig:Training_Schematic_Conditional}, we see that the conditional model yields very low errors, while the unconditional model has very large errors, especially for the largest $\sigma$ value, $\sigma=5$.
This is not surprising, as the image conditions (bottom right in Fig.\ \ref{fig:Training_Schematic_Conditional}) provide information about previous time steps, which provide a solid basis for the denoiser to predict the specific weather state we seek to predict, rather than a random weather state. Further, this method of adding the conditioning image as an input channel is similar to the conditioning method used in some Conditional Generative Adversarial Networks (CGANs) (e.g., \cite{Isola2017}).

Including the image condition, $\mathbf{c}$, in this manner can be easily expressed in terms of the original Karras framework \citep{Karras2022} by adding the condition, $\mathbf{c}$, to the score function, which becomes $\nablaxx \log p \big( \xx | \mathbf{c}; \sigma \big)$, and to the denoiser, which becomes $D_{\theta}(\xx | \mathbf{c};\sigma)$. 
The denoiser is then trained to minimize  

\begin{equation}
\label{eqn:expectation2}
\mathbb{E}_{\mathrm{c} \sim p_{\text{cond}}}\mathbb{E}_{\signal \sim \pdata} \mathbb{E}_{\noise \sim \mathcal{N}(\boldzero, \sigma^2 \boldi)} \lVert D_\theta(\signal + \noise | \mathrm{\mathbf{c}}; \sigma) - \signal \rVert^2_2,
\end{equation}
where $p_\text{cond}$ represents the distribution of the conditions used. In our case, $p_\text{cond} \approx \pdata \times \pdata$, because our condition consists of two images from the training dataset.
Once trained, the generation of new images is exactly the same as before, but now there is the condition passed along. One can think about the condition now informing the image generation path to a specific path in Fig.\ \ref{fig:image_generation_x_sigma}, where before the score could have pointed the image generation along any arbitrary path. 

This is not the only way to include a condition in diffusion models. \citet{Rozet2023}, \citet{Manshausen2024}, \citet{Qu_2024},  \citet{Martin2025} and \citet{Andry2025} break the score function into two tasks using Bayes rule: 
\begin{equation}
    \label{eqn:bayes_scorefn}
    \nablaxx \log p(\xx | \mathrm{\mathbf{c}}; \sigma) = \underbrace{\nablaxx \log p(\xx; \sigma)}_\text{Unconditional score} \\ + \underbrace{\nablaxx \log p(\mathrm{\mathbf{c}} | \xx; \sigma)}_\text{Guiding score}.
\end{equation}
This consists of the usual score function shown in Equation \ref{eq:prob_ode}, and a second part to be estimated by assimilating observations. This method of conditioning diffusion models is known in the computer science literature as \textit{guided diffusion}, where observations \textit{guide} the diffusion generation of new images. This alternative method has a lot of promise in data assimilation, but it is not yet clear how the temporal information we are using for forecasting would be incorporated into their framework. This is a topic of future research. 

\section{Data and Methods}
\label{sec:data_and_methods}

\subsection{Data}

The primary dataset of this paper is Channel 13 (10.3 $\mu m$ wavelength) from the the Advanced Baseline Imager \citep[ABI; ][]{schmit2017} on board the GOES-16 satellite, which was part of the Geostationary Operational Environmental Satellite (GOES) satellite series. Channel 13 is a longwave infrared channel and is chosen because of its wide use in general meteorology, particularly for detecting clouds at all times, including both day and night. GOES-16 was operational from December 2017 until April 2025 at a central longitude of -75 W.
Since the motivating goal of this paper is a global cloud forecast, we use full-disk (10 min resolution) data from April 2023 through June 2024. 

A full-disk image for Channel 13 has approximately two kilometer resolution near nadir (i.e., over the Amazon) and has a raw image shape of 5,424 by 5,424 pixels. Unfortunately, for the Graphical Processing Units (GPUs) available for this project (a single NVIDIA GH-200 system, hereafter GH-200) and the fixed underlying neural network architecture, the 5,424 by 5,424 image size is prohibitively large and involves gradient computations that would not fit into memory. Thus, we take an approach where we create our datasets by randomly choosing smaller patches of the larger image.

\subsubsection{Training set selection} 
The training set is selected from 2023 and includes several quality control steps. From all files in 2023, we select a random file with replacement. We check whether the previous and next time steps (minus and plus 10 mins) exist. If they exist, we open the files. From the 5,424 by 5,424 grid, we randomly sample 250 grid indices with a viewing zenith angle less than 65 degrees. This step avoids selecting imagery at the edge of the full disk where data are invalid. From these 250 grid incidences, we slice out a patch of 256 by 256 (about 500 km by 500 km, indicated as colored patches in Fig.\ \ref{fig:GOES_data_patches}). Out of these 250 patches we only keep those that have at least 10$\%$ cloud fraction. From the cloudy patches, we then include only patches that occur during the day (85 deg solar zenith). Daytime only data were chosen so auxiliary information like visible imagery and more accurate retrievals of cloud top height and base height are available for evaluation. From this final set of patches, we randomly choose 16 of them to save to avoid including too many samples from the same time step. We repeat this process until there are more than 30,000 patches for training. 

\subsubsection{Validation set selection} 
The same procedure is used to generate a validation set, but the files from January to August 2024 are used and only 1,000 total patches are used for evaluation. The primary use of this validation set is to determine the hyperparameters for image generation. 

\subsubsection{Test set selection}
The same procedure is used to generate a test set, but the files from August 2024 to February 2025 are used and only 1,000 total patches are used for evaluation.

\subsubsection{Preprocessing of images}
All images are scaled prior to training the diffusion models such that the mean of the training data is zero and the variance is one. While important for all neural networks, it is especially important for diffusion models because the variance scaling used for the Gaussian noise used during training and inference is highly dependent on the mean and range of the data. For example, if the noise is too small to replace the underlying data distribution, the diffusion models have a challenge generating new data from complete noise. The mean and standard deviation are calculated from the training dataset and applied to all three datasets. 

\subsection{Models Trained}

To explore the various avenues of diffusion for cloud nowcasting we train several different types of models. 
All models trained in this paper use the exact same underlying neural network architecture at their core, namely the U-Net architecture described below, but with different data sources. 

\subsubsection{U-Net architecture}

The U-Net choice follows an online tutorial available on HuggingFace (archived into the github repository for this paper, see Data Availability). It is a fairly large U-Net \citep{Ronneberger2015} with six up- and down- sampling blocks, which contain two res-net blocks \citep{He2015} each. There is also attention layers added at the fifth of the six total blocks and inside the bottleneck. This model has 113 million learnable parameters. We acknowledge that only using one core architecture limits the generalization of this paper, but due to long training times (order of days) on our GH-200, we are limited in our ability to perform hyperparameter tuning of the architecture. Using smaller core models (i.e., with fewer parameters) and core models without attention will likely speed up training and inference times for satellite nowcasting and is a topic of future research. 

\subsubsection{Baseline model (plain U-Net)}

To serve as a baseline forecast, we first train the plain U-Net in the more conventional manner on the training data. That is, we train the U-Net to perform the 10 min forecast using mean squared error as the loss. This \textit{plain U-Net} is trained using a random $80/20$ split of the training data, and allowed to train until early stopping when the held out validation loss stopped decreasing for 10 epochs. 

\subsubsection{Diffusion models}
We train three different diffusion variants that all use the U-Net as denoiser. The first diffusion model, named {\bf Diff} follows the \citet{Karras2022} framework, where the denoiser is denoising the 10 min forecast. The second diffusion model, named {\bf CorrDiff}, follows the approach from \citet{Mardani2025} and \citet{Pathak2024}. It uses the pre-trained U-Net of the baseline model to get an initial, potentially blurry, prediction of the 10 min forecast. Then it trains a denoiser to improve the forecast of the plain U-Net. The denoiser uses the same condition as Diff, along with the estimate obtained from the pre-trained U-Net output (i.e., the input is now 4 channels). The other change compared to Diff is that the CorrDiff model predicts the difference between the plain U-Net estimate and the truth. 

For all diffusion models, we train the denoisers for 1,000 epochs with a batch size of 45 and a two-batch gradient accumulation (i.e., effective batch size of 90). The networks then see more than 35 million images with various noise levels during training.
The Diff and CorrDiff diffusion models take about 5 days of training on our GH-200 with $98\%$ average GPU utilization. 

Given the relatively long training (evaluation) time for the Diff and CorrDiff models, we also explore a latent diffusion model \citep[e.g.,][]{Leinonen2023}, named {\bf LDM}. That is, we use a pre-trained variational autoencoder (VAE) from HuggingFace to compress our original data from 256 by 256 (with one channel) down to 64 by 64 pixels (with four channels). Due to the channel dimension going up, the compression factor is only four. The choice of VAE is discussed in Section \ref{sec:data_and_methods}\ref{sec:vaes}, but know that we choose the autoencoder that has the smallest reconstruction loss on the GOES imagery (Table 1). We then use the compressed data to train the denoiser for diffusion in this latent space. This results in a five-times speed-up in training, where the latent model completed training for 1,000 epochs in about 1 day. We attempted a latent CorrDiff model by training the corresponding U-Net model entirely in the latent space, but the U-Net predictions in the latent space were poor on their own and we decided to not pursue it further for this paper. 

\subsection{Off-the-Shelf Variational Auto Encoders}
\label{sec:vaes}

\begin{figure*}[t]
 \begin{center}
 \includegraphics[width=6in]{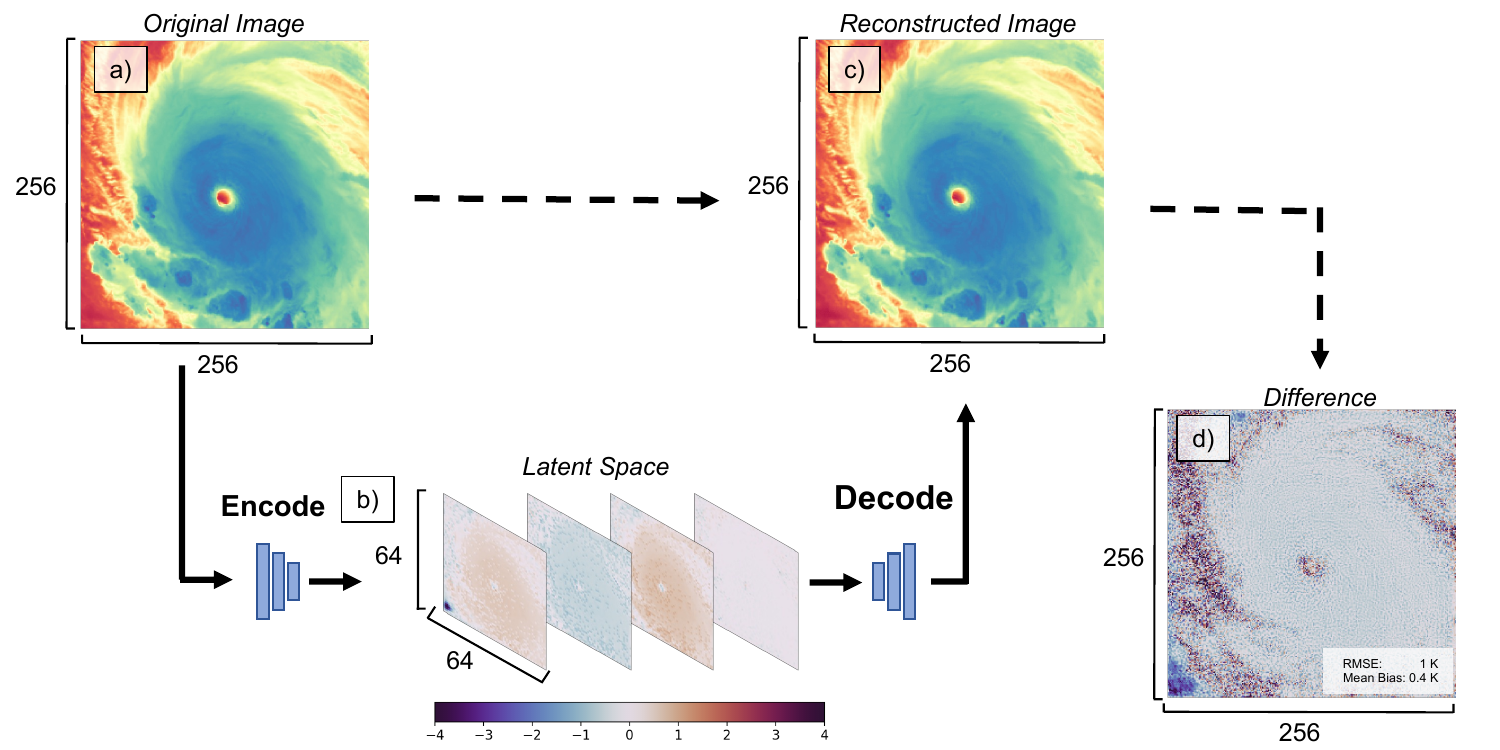}\\
 \caption{Example reconstruction of Hurricane Lee through the \textit{radames} pretrained VAE. (a) original IR brightness temperature image from Hurricane Lee. (b) The latent space where the single Hurricane Lee image has been compressed to 64 by 64 and have added 3 additional channels. (c) the reconstructed image of Hurricane Lee from the latent space. (d) The difference between (a) and (c).}
 \label{fig:VAE}
 \end{center}
\end{figure*}

The latent diffusion model requires the use of a variational autoencoder. Here we discuss the use of pre-trained variational autoencoders (VAE) for this purpose, specifically for compressing meteorological imagery. The availability of open-source pre-trained models have accelerated many efforts in the computer science field \citep[e.g., Stable Diffusion;][]{Rombach2022}. Given these pre-trained autoencoders are trained on pictures of cats, dogs, people, places and things, we were curious if these pre-trained models could be used accurately on meteorological satellite data. We tested eight total VAEs and the reconstruction losses for each one are found in Table \ref{tab:vae_comparison}. In general, most of them have relatively poor reconstructions, showing RMSEs of two and three kelvin where any individual pixel could be upwards of 20-30 K different than the original image. There is one that we tested that did have reconstruction losses less than 1 K. This VAE (example shown in Fig.\ \ref{fig:VAE}) was trained to perform super-resolution, and encodes the data spatially. Given the low reconstruction loss, we use this VAE for compressing our original satellite data from 256 by 256 pixels to 64 by 64 and training the diffusion in the 64 by 64 space (latent diffusion model). As will be seen later, while the latent diffusion can be skillful, it has some undesired qualities that might be a result of the VAE itself. Tuning the pre-trained VAE or training a VAE from scratch on satellite data specifically are both topics of future research. 

\begin{table*}
\begin{center}
\begin{tabular}{ |p{1.5cm}||p{3.5cm}||p{1.25cm}||p{1.25cm}||p{1.25cm}|}
\hline
\multicolumn{5}{|c|}{HuggingFace available VAE results on GOES Imagery} \\
\hline
User & Model name & Bias [K] & MAE [K] & RMSE [K] \\
\hline
CompVis & stable-diffusion-v1-4 &  1.48 & 8.81 & 2.97 \\
\hline
CompVis & ldm-super-resolution-4x-openimages & -0.64 & 3.02 & 1.74 \\
\hline
CompVis  & ldm-text2im-large-256 & 1.48 & 8.81 & 2.97 \\
\hline
stabilityai  & sdxl-vae & \textbf{-0.02} & 4.82 & 2.20 \\
\hline
stabilityai  & sd-vae-ft-mse  & -0.04 & 5.82 & 2.41 \\
\hline
Justin-Choo & epiCRealism-Natural-Sin-RC1-VAE & 0.10 & 8.42 & 2.90 \\
\hline
SG161222 & Realistic-Vision-V3.0-VAE & -0.04 & 5.82 & 2.41 \\
\hline
{\bf radames} & stable-diffusion-x4-upscaler-img2img  & -0.13 & \textbf{0.97} & \textbf{0.99} \\
\hline
\end{tabular}
\end{center}
\caption{A handful of available VAE models available on Huggingface used to quantify the reconstruction errors when run on the training dataset of GOES Imagery.}\label{tab:vae_comparison}
\end{table*}

\subsection{Generation Details and Ensembles}

As described in Section \ref{sec:intro_to_diff}, diffusion machine learning tasks are split into training the denoiser and generation. With this split come a few additional hyperparameters solely linked to the generation procedure. These generation hyperparameters are centered on: how many denoising steps to take (\texttt{num\_steps}); the maximum noise level to start from (\texttt{sigma\_max}); the spacing of the steps (\texttt{rho}); the amount of stochastic noise injected into the denoising process (\texttt{S\_churn}); and the scale of noise added back in (\texttt{S\_noise}). Just like training hyperparameters, a hyperparameter search would ideally be done for all diffusion models trained. Unfortunately, given the computational demand to run Diff and the CorrDiff model, we were only able to do a sparse hyperparameter search on the latent diffusion model and assume that these parameters work \textit{well} for the others. This assumption may be false, leading to suboptimal parameter sets, but this is an area of future research. More specifically, we perform a gridsearch of four of the above hyperparameters for the latent model and then choose the best joint performance of root mean squared error and the spread–skill ratio on the 1,000 example validation (i.e., use 10 different noisy images to start with the same two condition images). The specific values in the gridsearch are in Table \ref{tab:gen_hyperparameters}.

\begin{table}[htp]
\begin{center}
\begin{tabular}{ |p{3.5cm}||p{3.5cm}|}
\hline
\multicolumn{2}{|c|}{Generation Hyperparameters} \\
\hline
Hyperparameter & Value \\
\hline
\texttt{num\_steps}   &  [9, 18, \textbf{36}, 72]\\
\hline
\texttt{S\_churn}   & [0, \textbf{0.2}, 0.41421356237] \\
\hline
\texttt{sigma\_max} & [20, 80, \textbf{140}]\\
\hline
\texttt{rho}  &  [4, 7, \textbf{10}] \\
\hline
\end{tabular}
\caption{Grid search of generation hyperparameters for the diffusion models. The selected parameters are the bolded ones, heuristically selected from minimizing RMSE and the closest spread-skill ratio to 1. The \texttt{S\_churn} value is the \textit{effective} value; if using the code with this manuscript, be sure to multiply it by the \texttt{num\_steps} parameter.}
\label{tab:gen_hyperparameters}
\end{center}
\end{table}

Notice the mention of the spread-skill ratio, which is a metric for ensemble skill \citep{Haynes2023}. A benefit of the diffusion based models is the ability to generate as many samples as desired. This is partly controlled by the hyperparamters discussed above but also by varying the choice noisy image along the zeroth dimension of the input data to the neural network (see Fig.\ \ref{fig:Training_Schematic_Conditional} and Fig.\ \ref{fig:GOES_data_patches}b). For the ensembles analyzed here, we vary this noisy dimension with 10 different noise seeds to then get 10 different realizations of the same forecast. This is done for all three diffusion models. We choose 10 here as a trade off between the memory (time) required for running the forecast and having a decent ensemble size. For the Diff and the CorrDiff models it takes approximately 10 minutes to generate 10 members of a three hour forecast on a single NVIDIA GH-200. LDM takes about one minute to perform the same task. 

\subsection{Autoregression}

\begin{figure*}[t]
 \begin{center}
 \includegraphics[width=6in]{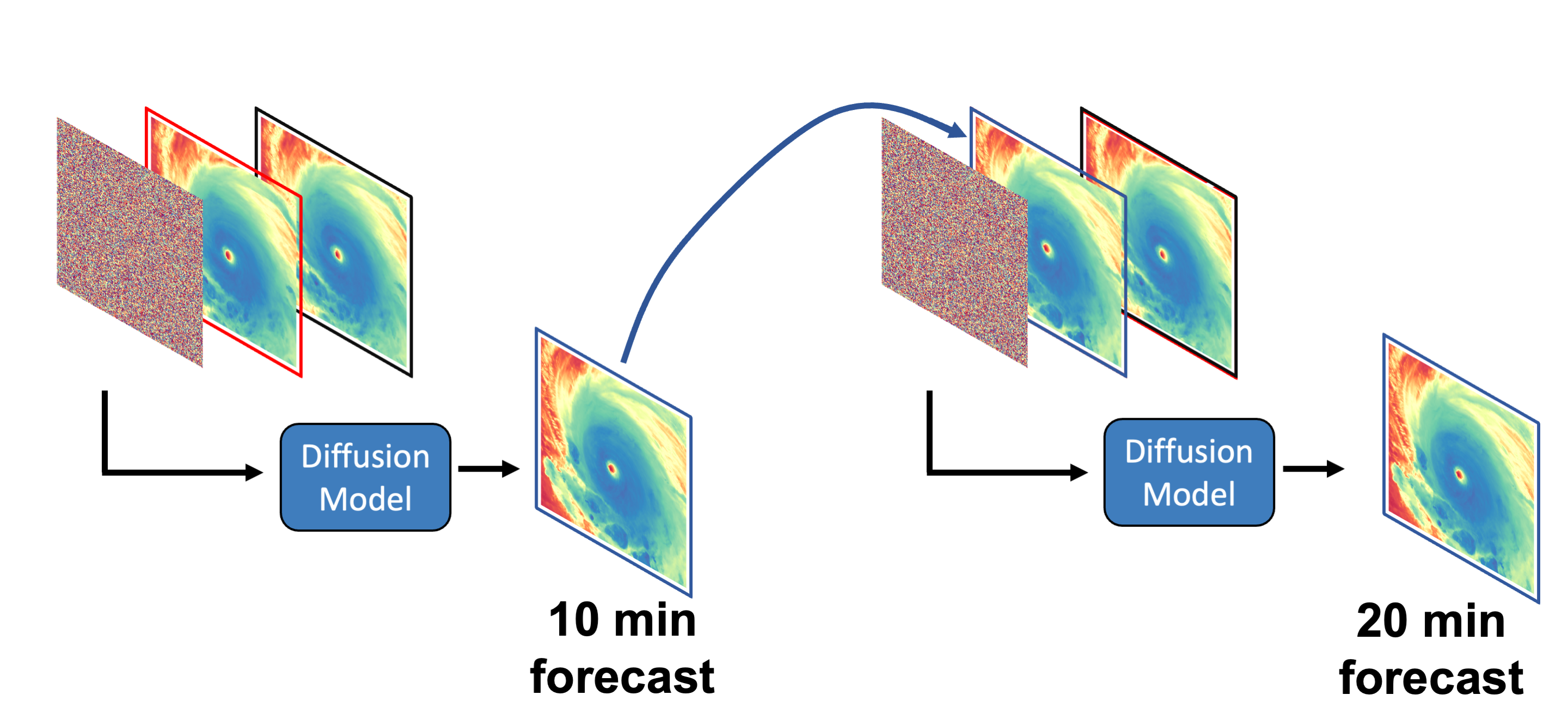}\\
 \caption{Graphic demonstrating the autoregressive task.}\label{fig:Autoregression}
 \end{center}
\end{figure*}

All machine learning forecast methods trained here follow the structure of many of the data-driven weather prediction models and optical flow techniques where two images ($\tau_0$, $\tau_{-10}$) are used to forecast a single time step ($\tau_{+10})$ (Fig.\ \ref{fig:GOES_data_patches}b and Fig.\ \ref{fig:Autoregression}). Then, the output from that single forecast can be fed back into the forecast models recursively. Forecasts of arbitrary length can then be rolled out. We choose to forecast out three hours (18 autoregressive steps). Some of the data driven models initially train on the single step prediction paradigm but then tune with a low learning rate on multiple steps. While technically possible to train diffusion through multiple inference steps, it would require very large amounts of VRAM not capable with a single GH-200.

\subsection{Metrics}
We choose a handful of common pixelwise metrics to get a general quantification of the forecast performance of the various methods. Those include mean error (ME), mean absolute error (MAE) and root mean square error (RMSE) all defined as: 

\begin{equation}
    \text{ME} = \frac{1}{n} \sum_{i=1}^{n} \frac{1}{m} \sum_{j=1}^{m}(y_{i, j} - \hat{y}_{i, j})
\end{equation}

\begin{equation}
    \text{MAE} = \frac{1}{n} \sum_{i=1}^{n} \frac{1}{m} \sum_{j=1}^m |y_{i, j} - \hat{y}_{i, j}|
\end{equation}

\begin{equation}
    \text{RMSE} = \sqrt{\frac{1}{n} \sum_{i=1}^{n} \frac{1}{m} \sum_{j=1}^m (y_{i,j} - \hat{y}_{i, j})^2}
\end{equation}

where $\hat{y}_{i, j}$ is the $j$th pixel in the $i$th forecasted image, $y_{i, j}$ is $j$th pixel in the $i$th true image, $n$ is the total number of images in the validation or testing set (which is 1000), and $m$ is the number of pixels per image (which is $256 \cdot 256$). Note that in the RMSE computation, we first compute MSE across the entire testing or validation set and then take the square root rather than computing RMSE for each image and then averaging those. We include mean error as a measure of bias (i.e., forecast drift).

\section{Results}
\label{sec:results}

We first present results for a case study to provide visual intuition for each model's characteristics, followed by performance metrics for the entire test set. The nowcast was built with the eventual use-case of forecasting cloud and precipitation location, so our evaluations focus on the evolution of cold brightness temperature regions in the imagery. 

\subsection{Case Study: The Coast of South America}

\begin{figure*}[t]
 \begin{center}
 \includegraphics[width=0.98\textwidth]{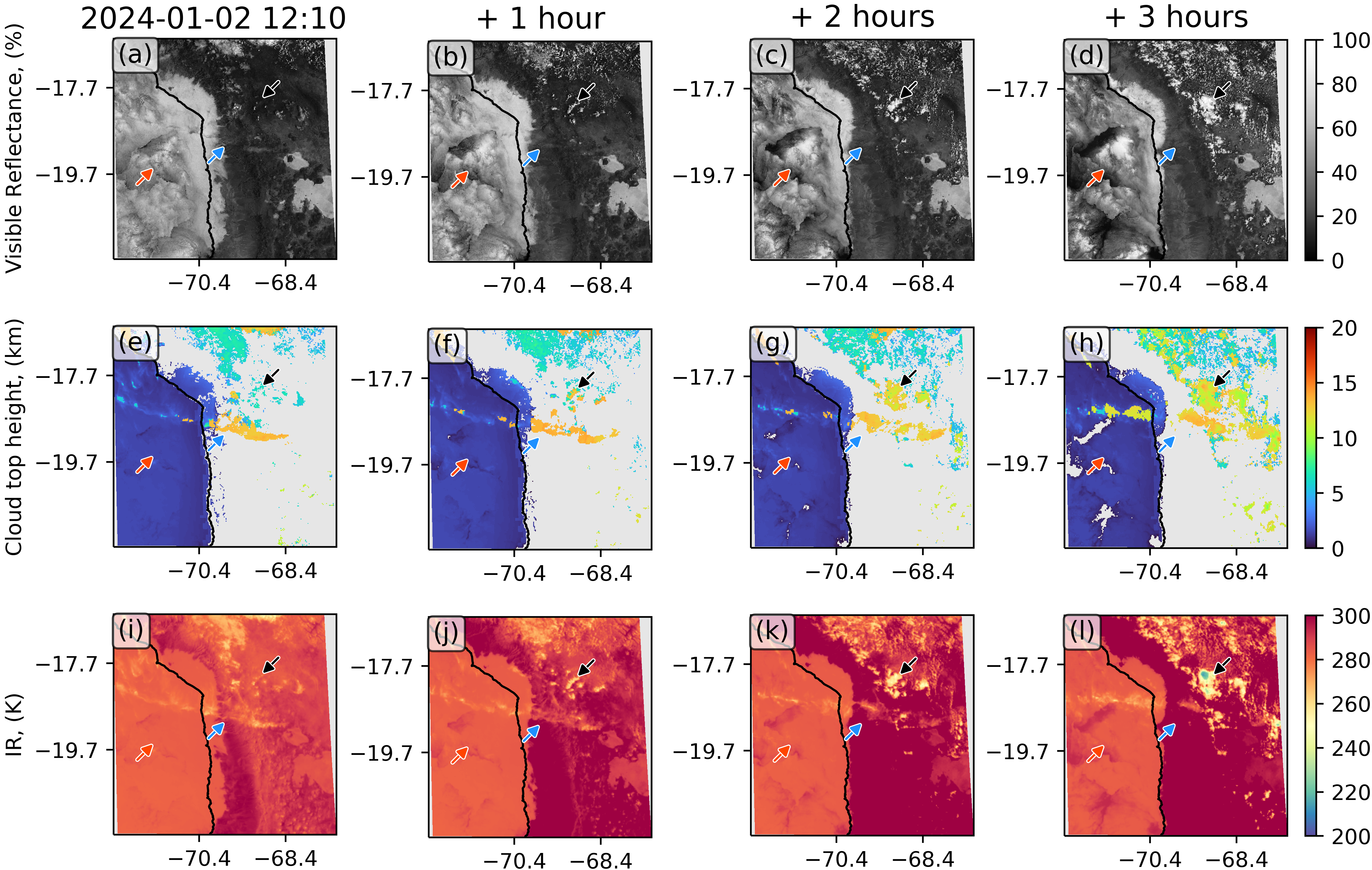}\\
 \caption{Context of the observed cloud evolution for a case from the west coast of South America. The evolution of visible reflectance (a - d), cloud top height in km (e-h), and infrared brightness temperature in K (i-l). Arrows mark features of interest, where the red arrow indicates stratocumulus, the blue arrow indicates cirrus clouds and the black arrow indicates a region of convection.}
 \label{fig:Case_Study_Setup}
 \end{center}
\end{figure*}
We provide an in depth case study that has a variety of cloud types to evaluate how well the various forecasts work with the gambit of clouds observed in the atmosphere. The case we show was chosen out of the validation set such that it contained low level and high level clouds, as well as clouds that form and decay over the three hour forecast. An overview of the case is shown in Fig.\ \ref{fig:Case_Study_Setup} showing the visible reflectance, the cloud top height retrieved from NOAA's Enterprise AWG Cloud Height Algorithm (ACHA; \citet{heidinger2018enterprise}), and the observed infrared brightness temperatures. 

For this case, there is a large stratocumulus deck extending from the Pacific Ocean onto the nearby land area of Chile (red arrow in Fig.\ \ref{fig:Case_Study_Setup}). There is also a streamer of cirrus cloud that is being advected into the domain from the west (blue arrow in Fig.\ \ref{fig:Case_Study_Setup}). Lastly, notice the development of some convection over the high terrain of the Andes as the atmosphere destabilizes through the afternoon (black arrow in Fig.\ \ref{fig:Case_Study_Setup}). 

\begin{figure*}[t]
 \begin{center}
 \includegraphics[width=0.8\textwidth]{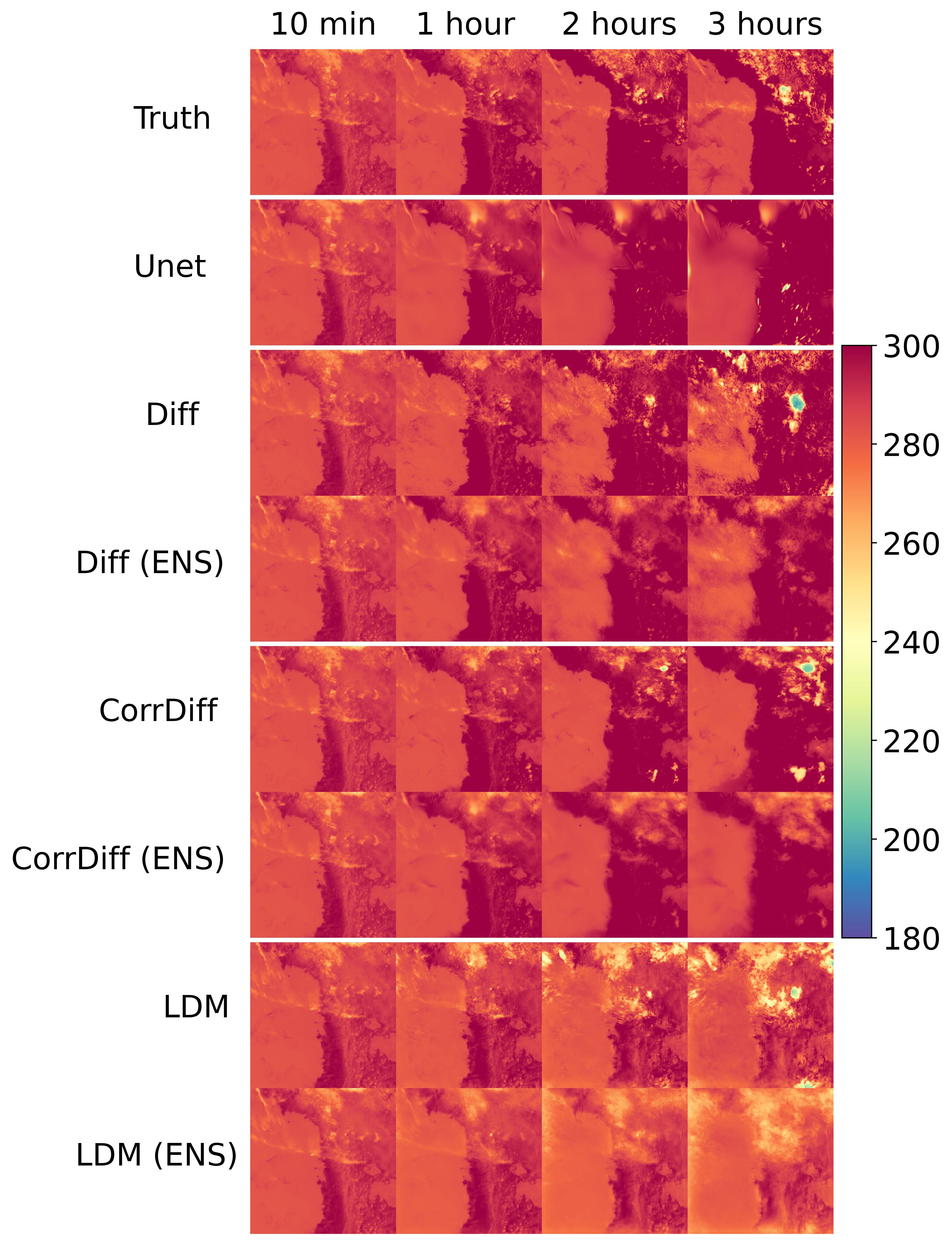}\\
 \caption{True and forecasted infrared brightness temperatures for the case in Fig.\ \ref{fig:Case_Study_Setup}. Across the rows are the truth, U-Net estimate, single member and 10-member ensemble mean of the Diff diffusion model, single member and 10-member ensemble mean of the CorrDiff diffusion model, and single member and 10-member ensemble mean of the LDM. The columns show different forecast times.}
 \label{fig:Case_Study_Forecast}
 \end{center}
\end{figure*}

Fig.\ \ref{fig:Case_Study_Forecast} contains all trained nowcast models along side the true infrared brightness temperatures for the same case. An animated version of this forecast can be found in the github repo associated with this paper (see Data Availability section). As expected, at a single 10 min forecast, all of the methods seem comparable from a qualitative sense (across the first row of Fig.\ \ref{fig:Case_Study_Forecast}).

\begin{figure*}[t]
 \begin{center}
 \includegraphics[width=0.8\textwidth]{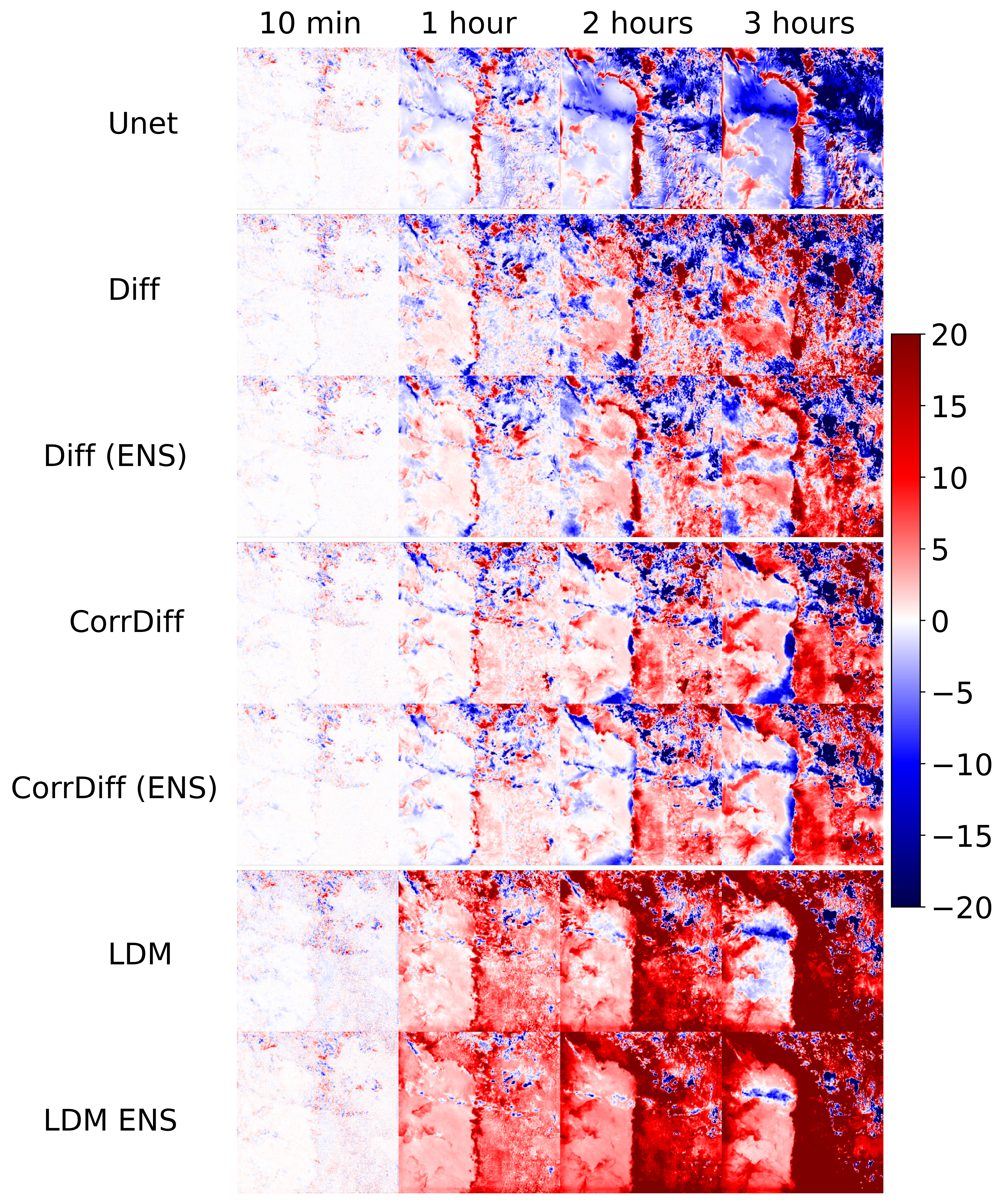}\\
 \caption{Difference plots for the forecasted infrared brightness temperatures shown in  Fig.\ \ref{fig:Case_Study_Forecast}. Namely, it is the truth row in Fig.\ \ref{fig:Case_Study_Forecast} (first row) minus each of the following rows.}
 \label{fig:Case_Study_Forecast_difference}
 \end{center}
\end{figure*}

For the one hour forecast, the blurry nature of the plain U-Net nowcast is discernible, the cirrus feature is soft and diffuse (second row in Fig. \ref{fig:Case_Study_Forecast}). Similarly, the stratocumulus deck to the west has become more uniform in its texture compared to the observed stratocumulus deck. These blurry features only get worse at longer lead times (two and three hours) and at the end of the forecast the cirrus cloud is missing and some potential artifacts are showing up in the NW corner of the image and over the land portion of the image to the SE. Looking at the difference between the U-Net and the truth (first row Fig. \ref{fig:Case_Study_Forecast_difference}), most of the domain has a warm bias that grows with lead time (blue hues increase with time). There is a cold bias where the stratocumulus deck doesn't erode as fast as reality along the coast. We are encouraged that the U-Net model does advect clouds initially, but after about one hour, the usefulness of the forecast seems limited. 

For the Diff approach, which uses the plain diffusion model in \citet{Karras2022}), the forecast loosely evolves like the observations. The stratocumulus deck recedes to the west, and has pieces of it break up. A surprising result was the ability of the diffusion model to initiate convection close to where convection was observed despite only knowing about 20 mins of observed satellite data. As a reminder, all models here have no information on the surface type, where the sun is, what time of day it is, etc. It is remarkable that the diffusion model can evolve a realistic looking scenario off of such limited information. Not all members of the diffusion model have convection in the same spot, evidenced by the ensemble mean being a bit smoother and no clear cold clouds like in the single member scenario. Clearly, the diffusion model is not perfect. The overall scene seems a bit noisy, suggesting the hyperparameters of the generation of forecasts are suboptimal. Furthermore, there is an overal warm bias of the scene, where the brightness temperature in the forecast are warmer than truth (blue hues in the third row of Fig. \ref{fig:Case_Study_Forecast_difference}). Overall we are encouraged by the ability of the diffusion model to evolve the scene and looking realistic (i.e., not blurry) in doing so out to three hours. 

For the CorrDiff approach, which uses the plain U-Net forecast as additional input, we can see that the scenes are considerably less noisy than for the Diff model (i.e., looks clearer). It could be that the hyperparameter set of the generation procedure is more optimal, or that the U-Net provides a good initial guess of the state and the diffusion can then just focus on small scale features. \citet{Mardani2025} suggest that the use of the U-Net helps reduce the variance of the task and makes it easier for the machine learning to learn. The CorrDiff model also convects the atmosphere like Diff model, but this time the convection shows up further to the NE than the observed convection. The stratocumulus retreat to the west is much better resolved by the CorrDiff model and does have a few breaks in the cloud deck. Also, as in the other forecast models, the cirrus streamer is not well captured beyond an hour, likely because of its tenuous nature. Similarly to the Diff model, the ensemble mean does not have a single cold cloud feature like the single member, but does have an area of colder clouds to the NE where cumulus and convection were observed. The CorrDiff model is not immune to the drift of the other forecast and infact overcorrects the U-Net input to now have a cold bias (red hues in the fourth row of Fig. \ref{fig:Case_Study_Forecast_difference}).

Considering the fastest of the diffusion models, the latent diffusion model (LDM), we can see for the first hour the evolution looks similar to the other diffusion models. After an hour, the edges of the image have a strong cold bias (red hues in the fifth row of Fig. \ref{fig:Case_Study_Forecast_difference}). We suspect this is because of the non-optimal encoding of the satellite images with the pre-trained autoencoder. If you consider Fig.\ \ref{fig:VAE}b, the edge of the image has some odd cold/hot pixels. This was common when inspecting the latent space (not shown) and leads to large errors in the encoding-decoding process (Fig.\ \ref{fig:VAE}d). The evolution of the surface warming also looks off compared to the other diffusion models and the observed data. Thus, while the LDM model can work, showing an evolution of clouds and even initiating convection, there are clear tradeoffs for its ability to run five to ten times faster. Furthermore, as mentioned before, the VAE used here is likely not yet optimal. 

\begin{figure}[t]
 \begin{center}
 \includegraphics[width=3in]{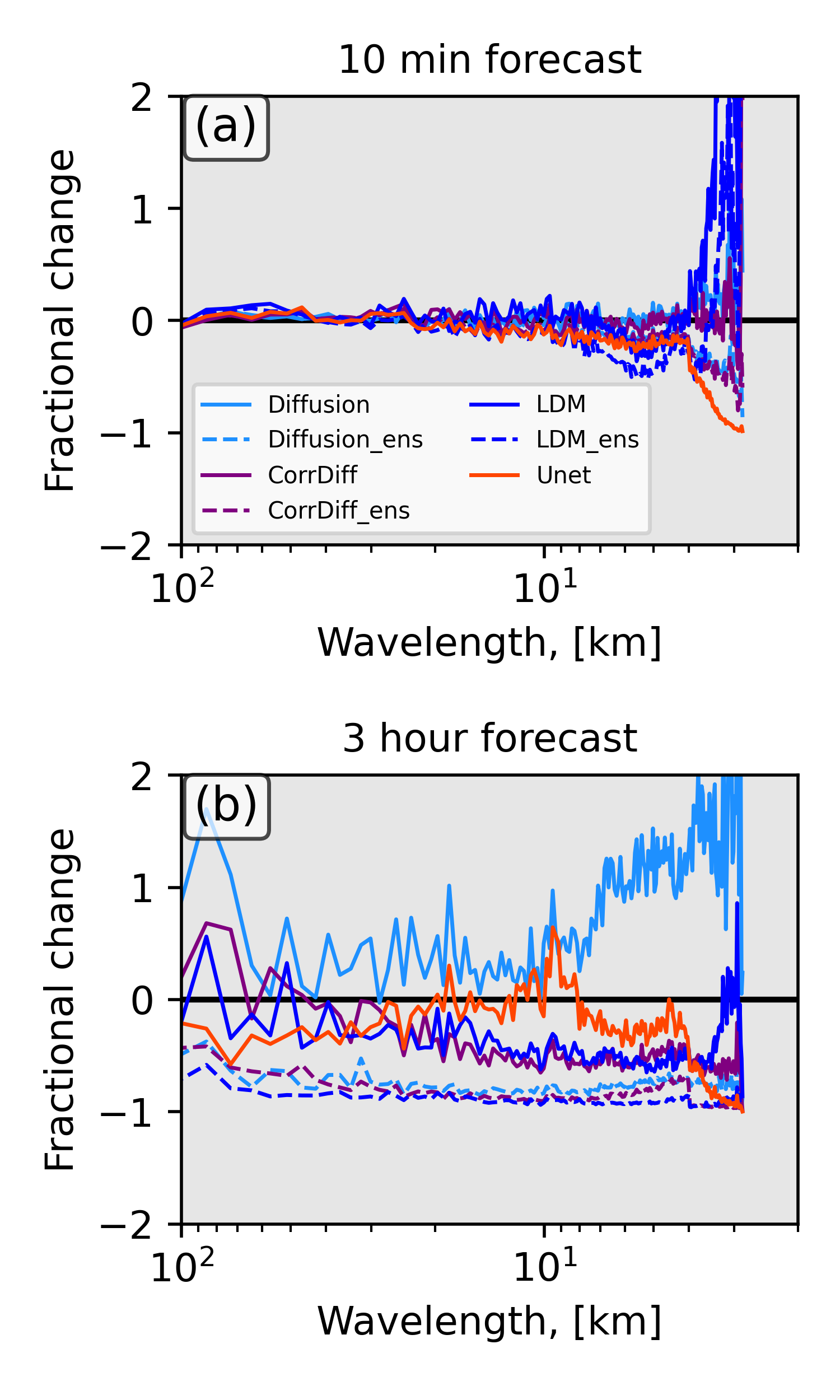}\\
 \caption{A measure of image sharpness, fractional powerspectra change for the images in (Fig.\ \ref{fig:Case_Study_Forecast}). The light blue line is the Diff forecast, purple line is CorrDiff, dark blue is the LDM forecast and the plain U-Net is in orange. The dashed lines are the ensemble means for each of the corresponding forecasts. (a) the 10 min forecast (b) the 3 hour forecast.}
 \label{fig:Case_Study_Spectra}
 \end{center}
\end{figure}

A motivation of this paper was to enhance the often blurry output of neural networks that optimize some variant of the MSE loss. To quantify this a bit further, we consider the power spectra of each nowcast for the same case and compare it to the original image. Fig.\ \ref{fig:Case_Study_Spectra} shows the fractional change for the 10 min forecast and the three hour forecast (Fig.\ \ref{fig:Case_Study_Forecast}). For the 10 min forecast, for wavelengths greater than 10 km all match well. For smaller features, the single member diffusion (solid light blue) and single member CorrDiff (solid purple) have similar or more of the original features than the U-Net and the ensemble means (dashed lines). While the usefulness of evaluating the sharpness of the ensemble mean is questionable, we include it here to provide a comparison point for how blurry the individual members could get. For waves smaller than 4 km (two times the native resolution) the U-Net effectively removes these features despite the image in Fig.\ \ref{fig:Case_Study_Forecast} looking fairly sharp. Similarly the ensemble means reduce the number of small waves (less than five kilometers), but not to the extent of the U-Net and this smoothing is expected given there could be divergence among the ensemble members. Surprisingly, the LDM (dark blue line) and to a less extent the Diff model (light blue line) add small scale features undetectable to the human eye in Fig.\ \ref{fig:Case_Study_Forecast} but more noticeable in the the difference plot with red and blue pixels being located next to each other across the image (\ref{fig:Case_Study_Forecast_difference}). These small wave artifacts in the LDM model are from the VAE itself, which was found to add noise in the decoding process (not shown). Meanwhile, we hypothesize that the small wave artifacts in the Diff model are left over from the denoising process of the diffusion. 

At the three hour forecast, the spectra are a bit harder to interpret because of how different the forecasts are from one another but the changes to the smoothness in the imagery is clear. The Diff model has more waves at all wavelengths, which is the noisy feature we described in Fig.\ \ref{fig:Case_Study_Forecast}. The ensemble means are all smoother than the U-Net, expected from the spread of the ensemble members. Even the CorrDiff model (solid purple line) shows some blurriness overall compared to the observed image, but does have larger spectra for waves smaller than four km, compared to the U-Net. Considering the imagery in \ref{fig:Case_Study_Forecast}, the blurriness of the CorrDiff model manifests itself as softer edges to the area of developing convection. As for the LDM at 3 hours, there are some shorter wavelength information that matches the original image and some loss of waves at middle wavelengths, but the growth of artifacts near the edge is likely impacting the interpretation of the spectra. For future research we plan to use additional sharpness metrics to better quantify the impact of diffusion on overall sharpness of the machine learning output 
\citep{sharpness2025}.

\subsection{Bulk Results}

\begin{figure*}[t]
 \begin{center}
 \includegraphics[width=0.98\textwidth]{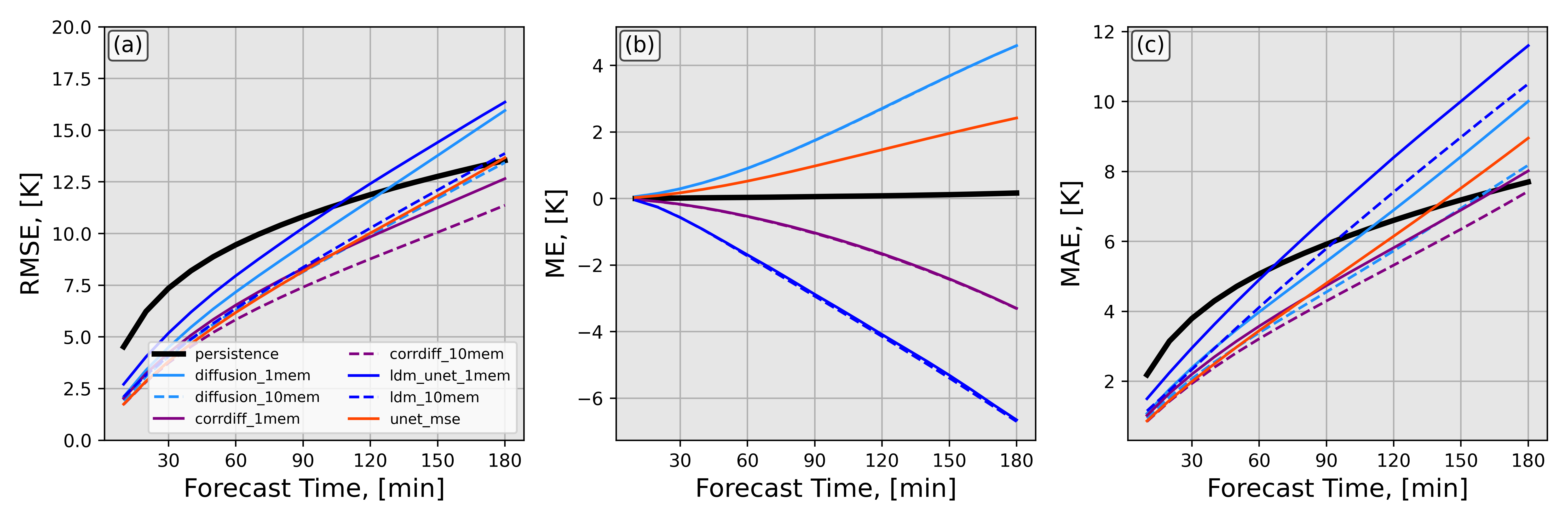}\\
 \caption{Bulk pixel-based metrics for 1,000 example scenes out of the validation set. (a) the root mean squared error between various forecasts and the true observation. The solid black line is a persistence forecast from $t_0$. The light blue solid line is the best diffusion model member of the 10 total members. The dashed light blue line is the 10 member diffusion model mean. Similarly, the dark blue lines are for the latent diffusion model and the purple lines are the CorrDiff models. The red solid line is the U-Net forecast. (b) same as (a) but for mean error. (c) same as (a) and (b) but the mean absolute error.}
 \label{fig:Bulk_Pixel_Metrics}
 \end{center}
\end{figure*}

To quantify broader performance we look at several pixel-based metrics across the 1,000 test set forecasts. As an additional baseline besides the U-Net, we also include persistence (i.e., the image at $\tau_0$). The best performing model according to RMSE is the mean of the 10-member CorrDiff ensemble, outperforming all other methods for the three hour forecast by about one kelvin (Fig.\ \ref{fig:Bulk_Pixel_Metrics}a) and the baseline forecasts (i.e., U-Net and persistence) by two kelvin. The next best performance is the single member CorrDiff model followed closely by the 10-member mean Diff model and then the U-Net and the mean of the 10-member latent diffusion model. Most methods outperform persistence for the full three hours except the single member latent diffusion model and the single member diffusion model. 

In terms of mean error (ME), all machine learning forecast methods have a bias. The latent diffusion models drift to cold values exceeding six kelvin by the end of the three hour forecast. The CorrDiff forecasts also drift cold, but only to about three kelvin at the end of the three hour forecast. The other two machine learning forecasts have a warm bias, where the U-Net tends to warm about two kelvin and the diffusion models drift to more than four kelvin. As for mean absolute error (MAE), only three of forecasts can perform similarly or outperform the persistence forecast. The 10-member mean ensemble of CorrDiff again is the best performer followed by the single member CorrDiff and then the 10-member mean diffusion model. 

We compare here mainly to persistence and the U-Net forecast. Fig.\ \ref{fig:Bulk_Pixel_Metrics} confirms what was shown qualitatively in Fig.\ \ref{fig:Case_Study_Forecast} where most of the diffusion models have added value over the more traditional U-Net approach and a simple persistence forecast. Furthermore, all methods not only advect clouds \citep[e.g., an optical flow solution;][]{Shakya2019}, they can also create and decay clouds. We are encouraged that these models work given the limited information content included as inputs. 

\begin{figure}[t]
 \begin{center}
 \includegraphics[width=3in]{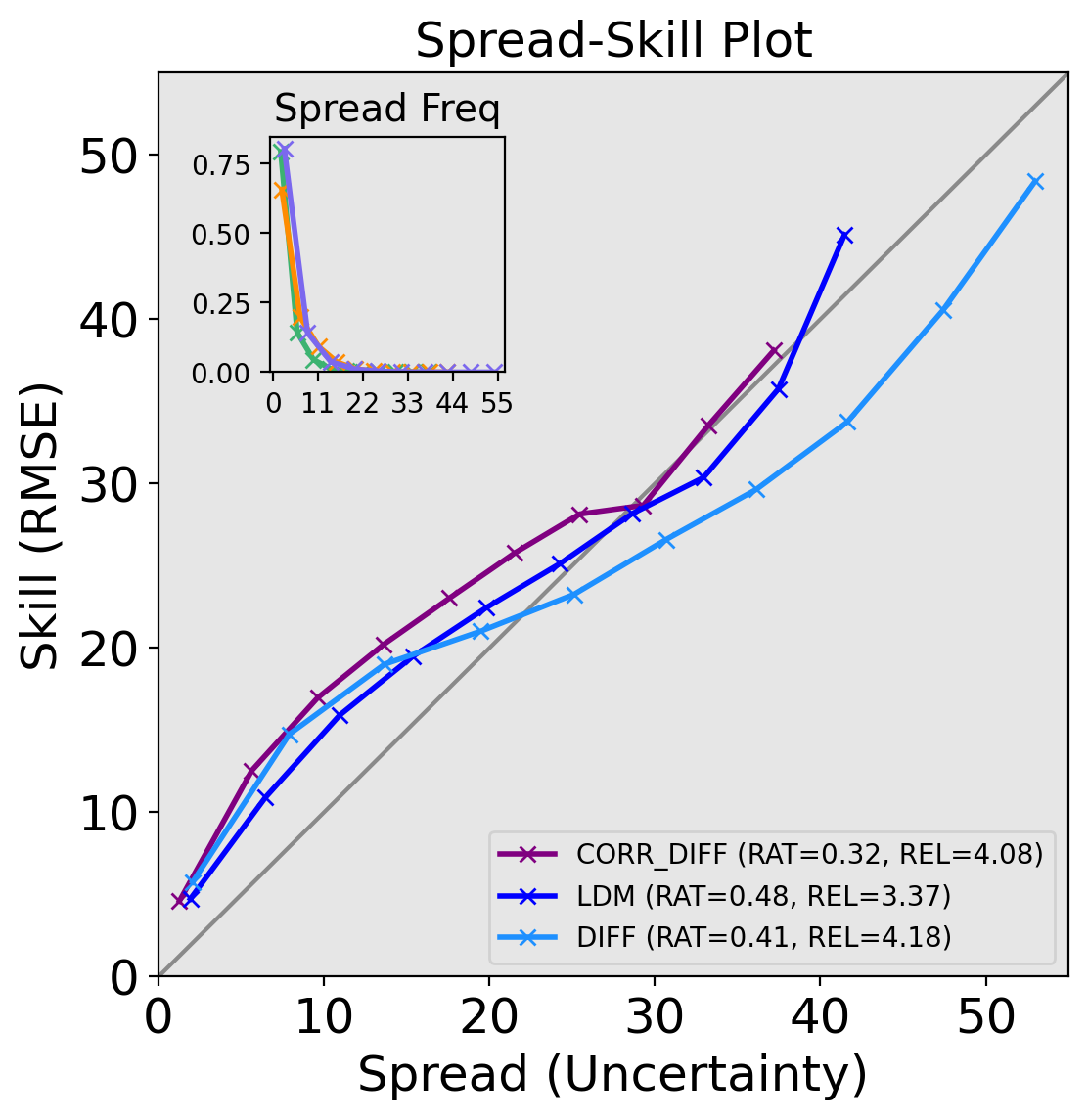}\\
 \caption{A spread-skill plot to describe the reliability of the uncertainty quantification of the various diffusion ensembles.}
 \label{fig:Spread_Skill}
 \end{center}
\end{figure}

A key perk that we have seen already in the diffusion framework is the ability to generate n-many samples to use ensemble based forecasts. This already results in improved pixel-based metrics in Fig \ref{fig:Bulk_Pixel_Metrics}. We can also look to quantify how well the uncertainty of that ensemble corresponds to the true error of the model. We leverage the skill-spread diagram \citep[e.g., ][]{Haynes2023, dellemonache2013probabilistic} where the standard deviation of the 10 member diffusion ensembles is compared to the RMSE of the ensemble (Fig.\ \ref{fig:Spread_Skill}). 

We see that for regions of smaller errors (e.g., RMSE less than 25-30 K), the 10-member ensembles are under-dispersive. This can be expected to an extent given the relatively small ensemble size of 10 members. Then as the error grows, the CorrDiff and the latent model are better calibrated showing the spread similar to the RMSE. As for the Diff model, it switches from under-dispersive to over-dispersive. Thus using the ensemble spread as a tool for this specific diffusion model to approximate forecast uncertainty becomes convoluted. An area of future research is to isolate if the ensembles are under-dispersive/over-dispersive for specific weather regimes (e.g., convection, frontal). This would help in the overall uncertainty quantification for the end user. 

%
%

\section{Discussion}
\label{sec:discussion}

\subsection{Lessons Learned}
As seen in the prior subsections, developing diffusion models requires numerous design choices to navigate the trade-offs of practical applications. 
Here we summarize and discuss some of these choices and trade-offs: 
\begin{itemize}
\item 
    We chose the overall framework of score-based diffusion models over DDPMs because they are significantly faster without presenting any disadvantages.
    
\item 
    The underlying architecture, while arbitrary, does have an affect on the timeliness of the model. As we were learning to use diffusion we neglected to change the underlying architecture from an online tutorial. Turns out this architecture, while skillful, had many parameters and some attention layers that made training and inference times considerably large (5 days to train; 10 minutes to run a 3 hour forecast for 10 ensemble members). This resulted in us exploring latent diffusion models to try to speed up the model. An alternative route would have been to investigate simpler underlying architectures that are faster (e.g., less parameters, careful placement of attention layers), or explore other community provided and optimized networks like NVIDIA's PhysicsNeMo which is used in \citet{Mardani2025} and \citet{Pathak2024}.

\item 
    Whether a latent model (LDM) should be used is a trade-off between higher inference speed and potentially lower accuracy. In our experience, conducting experiments with and without latent layers is the only way to tell whether they are worth using for a specific application. Experiments should include trying different latent space dimensions and different types of VAEs inside the LDM. A challenge is the large space of hyperparameters to explore in the LDM space, especially given that diffusion models are generally slow to train. Another challenge is training a VAE that faithfully reconstructs the training data. That is, any artifacts (blurring, added noise, token artifacts etc) created by the VAE are not correctable from the diffusion model and will end up in the final output.
\item 
    An even bigger question for any specific application is whether diffusion models are worth the effort, including more complex code base and large computational costs for both training and inference. For example, ECMWF explored two options to extend AIFS - their AI-based global weather forecasting model \citep[][]{Lang2024a} - to provide probabilistic forecasts: (1) integrating diffusion models \citep[]{Chantry2025AIFS}, or (2) utilizing the Continuous Ranked Probability Score (CRPS) \citep[][]{Lang2024b} as a loss function to be used with a simpler neural network. After trying both approaches the CRPS route was chosen for first operationalization. Similarly, Google's operational weather forecast models have seen a shift from Diffusion with their GenCast model \citep{Price2025} to a similar CRPS loss with their FGN model \citep{Alet2025}, where the FGN model is now in operations. Following these two efforts, other global weather models have also used CRPS losses to get to ensemble forecasting \citep[e.g., FuXi-ENS;][]{Zhong2025}, suggesting that those implementing weather forecasts have been focusing on implementation speed over potentially sharper forecasts with diffusion. Even though diffusion is currently costly in training and inference, there are many researching how to speed up diffusion \citep[e.g.,][]{Xu2025}. 
\end{itemize}

\subsection{Limitations and Future Directions}

The diffusion models we showed here are far from perfect. The Diff model tends to be noisier than the CorrDiff approach, which could be detrimental to longer forecast lead times. Furthermore, all machine learning methods had a clear bias with time, some trending warm, others trending cold, which could eventually lead to nonphysical forecasts. Another challenge of the diffusion models is their increased computational cost compared to the more traditional U-Net (i.e., MSE loss). It took about five days to train the plain diffusion model and the CorrDiff model on a relatively new GPU system (e.g., GH-200). This length of training time is prohibitively long to allow for a hyperparameter search of the underlying architecture on our resources. Furthermore, unlike in more traditional machine learning methods, the computational cost is not isolated to training time. Because of the iterative steps of denoising, the diffusion models take substantially longer to run during inference time, 10 mins to generate a three hour forecast with 10-members for the diffusion model versus the 10 seconds for the U-Net. This added computational time also prohibited hyperparameter tuning of \textit{generation} parameters. Thus, the choices we made are likely sub-optimal. We also explored the use of pretrained autoencoders to enable the use of latent diffusion, which substantially improved computation costs (one day to train, one min to run forecast), but results are clearly worse than for the other diffusion models. The LDM approach might improve if a specialized autoencoder is trained for satellite imagery. 

Since we were compute constrained, we only had computational time to investigate a single common neural network architecture across all experiments. This is a key limitation of this paper. Other architectures should have been explored, including simpler ones (e.g., a U-Net with less parameters or no attention layers) that could have sped up the training and inference speed as well as more advanced networks (e.g., vision transformers and graphs). Another limitation of this work was the lack of comparison of the trained machine learning methods here to other methods of forecasting satellite brightness temperatures. Future work would ideally include forecasts from both traditional numerical weather prediction \citep[e.g.,][]{Griffin2017,Griffin2024} and optical flow \citep[e.g.,][]{Shakya2019}. Even though the diffusion models trained here have considerable compute costs, they are likely faster than traditional NWP thus providing added benefit. A clearly outlined study comparing the three would be greatly beneficial to the community and is a topic of future research.

This paper closely followed the methodology from \citet{Karras2022} which was also used in \citet{Mardani2025}, \cite{Manshausen2024} and \citet{Price2025}. Since the original \citet{Karras2022} work, there have been improvements. The improvements are mainly centered on carefully choosing the machine learning architecture \citep{Karras2024a} and having better image generation \citep{Karras2024b}. Thus, there are improvements to the methodology directly that could improve results here.

Outside of changes to the methodology, other ways to improve the satellite nowcasting would be to include additional information (i.e., more conditions) to the model input. Specifically, including simple input fields like the solar zenith angle and topography should help inform the model where in the image should be warming and what the surface type is (e.g., land vs.\ ocean), respectively. One could simplify the task further by breaking the task into two parts, advection and alteration. Advection could be handled well by known tools like optical flow and the diffusion model could handle the generation, growth, and/or decay of clouds. In addition to the solar zenith angle and the topography, this work could be extended to include more channels of satellite imagery. This idea of multitask learning might also help because if a multi-channel diffusion model was trained using all thermal bands, then the model could potentially learn more of the 3-dimensional structure of the atmosphere, as well as the water vapor content, which are both critical for the evolution of clouds.

\section{Conclusions}
\label{sec:conclusions}

A challenge with many machine learning methods is the prevalence of blurry solutions as a result of mean squared error losses that then have limited usefulness in human-in-the-loop applications (i.e., weather forecasters). In the literature, diffusion models suggest to be a solution to these blurry outputs while also providing useful uncertainty quantification through ensembles. For this paper, we delve into the intuition behind the success of diffusion models for providing sharper images, we explore the use of score-based diffusion models for a basic forecasting task (in our case to nowcast geostationary satellite imagery), and we highlight lessons learned from our efforts for both research applications and operational tasks. 

This work has three main contributions aimed towards aiding the community in adopting diffusion modeling. First, we provide an intuitive introduction to the methodology of score-based diffusion models, discuss how to train and use score-based diffusion models, and detail often overlooked steps needed for earth science applications that are not addressed in the computer science literature (Section \ref{sec:intro_to_diff}). Second, we explore various types of score-based diffusion models and evaluate how they perform for our nowcasting task of forecasting GOES channel 13 infrared brightness temperatures out to three hours. More specifically, we explore three different diffusion modeling approaches: a standard score-based diffusion model (Diff) that follows \citet{Karras2022}, a residual corrective diffusion model (CorrDiff) that follows \citet{Mardani2025}, and a latent diffusion model (LDM) inspired by \citet{Leinonen2023}. We showed that the diffusion models were able to nowcast satellite imagery qualitatively and quantitatively better than both persistence forecasts and a mean-squared-error-trained U-Net (Figures 10-14). The best performing diffusion model was the CorrDiff approach followed by the Diff model and then the LDM. All diffusion models were able to not only advect the current clouds in the scene but also generate and decay clouds, a key weakness of optical flow techniques. We also showed that all methods were able to generate convection (i.e., localized regions of very cold brightness temperatures) despite only having two images (20 mins) of 10 micron infrared imagery and no information on the thermodynamic state of the atmosphere, the solar input, or the land surface type. Third, we provide valuable perspectives on utilizing diffusion models both for research and operationally, discussing trade-offs that need to be considered between accuracy and computing resources (Section \ref{sec:discussion}).
 
Overall, we are encouraged by the prospect of score-based diffusion models for machine learning tasks in the Earth System sciences. They provide often high-resolution features that appear more realistic than the more traditional U-Net approaches. Furthermore, the out-of-the-box uncertainty quantification offered from ensemble generation is an added benefit, often improving pixel-wise metrics while still having the information content of the individual members. We hope that this work provides intuition on how diffusion modeling works as well as valuable tips to accelerating their use in the Earth System sciences.

\acknowledgments
We thank the three reviewers and the editor for their comments and guidance in improving this paper. The work of all authors was supported by the Office of Naval Research as part of the OVERCAST project under award N0001424C2214. We would like to thank our colleagues, John Haynes, Kyle Hillburn, Jebb Stewart, Marie McGraw, Ryan Lagerquist, Yoonjin Lee, Chuck White, and Marshall Baldwin from CIRA, and Alex Rybchuk from NREL, for thought provoking discussions and preliminary feedback on this paper over the past couple years.  
%
%
\datastatement
The code associated with this manuscript can be found on GITHUB at \url{https://github.com/dopplerchase/cira-diff}. Additionally, the code will be archived alongside the validation dataset of this manuscript on Dryad at the time of publication. 

%





%



\bibliographystyle{ametsocV6}
\bibliography{references}

@STRING{AN        = "Astrophys.\ Norv."}

@STRING{MA        = "Meteor.\ Appl."}

@STRING{OCEAN     = "Oceanography"}

@inproceedings{Karras2022,
 author = {Karras, Tero and Aittala, Miika and Aila, Timo and Laine, Samuli},
 booktitle = {Advances in Neural Information Processing Systems},
 editor = {S. Koyejo and S. Mohamed and A. Agarwal and D. Belgrave and K. Cho and A. Oh},
 pages = {26565--26577},
 publisher = {Curran Associates, Inc.},
 title = {Elucidating the Design Space of Diffusion-Based Generative Models},
 url = {https://proceedings.neurips.cc/paper_files/paper/2022/file/a98846e9d9cc01cfb87eb694d946ce6b-Paper-Conference.pdf},
 volume = {35},
 year = {2022}
}

@article {Chase2022,
      author = "Randy J. Chase and David R. Harrison and Amanda Burke and Gary M. Lackmann and Amy McGovern",
      title = "A Machine Learning Tutorial for Operational Meteorology. Part {I}: Traditional Machine Learning",
      journal = "Weather and Forecasting",
      year = "2022",
      publisher = "American Meteorological Society",
      address = "Boston MA, USA",
      volume = "37",
      number = "8",
      doi = "10.1175/WAF-D-22-0070.1",
      pages=      "1509 - 1529",
      url = "https://journals.ametsoc.org/view/journals/wefo/37/8/WAF-D-22-0070.1.xml"
}

@misc{Song2021,
      title={Score-Based Generative Modeling through Stochastic Differential Equations}, 
      author={Yang Song and Jascha Sohl-Dickstein and Diederik P. Kingma and Abhishek Kumar and Stefano Ermon and Ben Poole},
      year={2021},
      eprint={2011.13456},
      archivePrefix={arXiv},
      primaryClass={cs.LG},
      url={https://arxiv.org/abs/2011.13456}, 
}

@article{ho2020,
  title={Denoising diffusion probabilistic models},
  author={Ho, Jonathan and Jain, Ajay and Abbeel, Pieter},
  journal={Advances in neural information processing systems},
  volume={33},
  pages={6840--6851},
  year={2020}
}

@misc{Manshausen2024,
      title={Generative Data Assimilation of Sparse Weather Station Observations at Kilometer Scales}, 
      author={Peter Manshausen and Yair Cohen and Jaideep Pathak and Mike Pritchard and Piyush Garg and Morteza Mardani and Karthik Kashinath and Simon Byrne and Noah Brenowitz},
      year={2024},
      eprint={2406.16947},
      archivePrefix={arXiv},
      primaryClass={cs.LG},
      url={https://arxiv.org/abs/2406.16947}, 
}

@inproceedings{Rozet2023,
 author = {Rozet, Fran\c{c}ois and Louppe, Gilles},
 booktitle = {Advances in Neural Information Processing Systems},
 editor = {A. Oh and T. Naumann and A. Globerson and K. Saenko and M. Hardt and S. Levine},
 pages = {40521--40541},
 publisher = {Curran Associates, Inc.},
 title = {Score-based Data Assimilation},
 url = {https://proceedings.neurips.cc/paper_files/paper/2023/file/7f7fa581cc8a1970a4332920cdf87395-Paper-Conference.pdf},
 volume = {36},
 year = {2023}
}

@article{Mardani2025,
  author = {Mardani, Morteza and Brenowitz, Noah and Cohen, Yair and Pathak, Jaideep and Chen, Chieh-Yu and Liu, Cheng-Chin and Vahdat, Arash and Nabian, Mohammad Amin and Ge, Tao and Subramaniam, Akshay and Kashinath, Karthik and Kautz, Jan and Pritchard, Mike},
  year = {2025},
  month = {February},
  title = {Residual Corrective Diffusion Modeling for km-Scale Atmospheric Downscaling},
  journal = {Communications Earth \& Environment},
  volume = {6},
  number = {1},
  pages = {124},
  doi = {10.1038/s43247-025-02042-5},
  url = {https://doi.org/10.1038/s43247-025-02042-5}
}

@misc{Pathak2024,
      title={Kilometer-Scale Convection Allowing Model Emulation using Generative Diffusion Modeling}, 
      author={Jaideep Pathak and Yair Cohen and Piyush Garg and Peter Harrington and Noah Brenowitz and Dale Durran and Morteza Mardani and Arash Vahdat and Shaoming Xu and Karthik Kashinath and Michael Pritchard},
      year={2024},
      eprint={2408.10958},
      archivePrefix={arXiv},
      primaryClass={physics.ao-ph},
      url={https://arxiv.org/abs/2408.10958}, 
}

@misc{Leinonen2023,
      title={Latent diffusion models for generative precipitation nowcasting with accurate uncertainty quantification}, 
      author={Jussi Leinonen and Ulrich Hamann and Daniele Nerini and Urs Germann and Gabriele Franch},
      year={2023},
      eprint={2304.12891},
      archivePrefix={arXiv},
      primaryClass={physics.ao-ph},
      url={https://arxiv.org/abs/2304.12891}, 
}

@article{Price2025,
  author = {Price, Ilan and Sanchez-Gonzalez, Alvaro and Alet, Ferran and Andersson, Tom R. and El-Kadi, Andrew and Masters, Dominic and Ewalds, Timo and Stott, Jacklynn and Mohamed, Shakir and Battaglia, Peter and Lam, Remi and Willson, Matthew},
  title = {Probabilistic weather forecasting with machine learning},
  journal = {Nature},
  year = {2025},
  volume = {637},
  number = {8044},
  pages = {84--90},
  doi = {10.1038/s41586-024-08252-9},
  url = {https://doi.org/10.1038/s41586-024-08252-9}
}

@misc{Karras2024a,
      title={Analyzing and Improving the Training Dynamics of Diffusion Models}, 
      author={Tero Karras and Miika Aittala and Jaakko Lehtinen and Janne Hellsten and Timo Aila and Samuli Laine},
      year={2024},
      eprint={2312.02696},
      archivePrefix={arXiv},
      primaryClass={cs.CV},
      url={https://arxiv.org/abs/2312.02696}, 
}

@misc{Karras2024b,
      title={Guiding a Diffusion Model with a Bad Version of Itself}, 
      author={Tero Karras and Miika Aittala and Tuomas Kynkäänniemi and Jaakko Lehtinen and Timo Aila and Samuli Laine},
      year={2024},
      eprint={2406.02507},
      archivePrefix={arXiv},
      primaryClass={cs.CV},
      url={https://arxiv.org/abs/2406.02507}, 
}

@article{Bowler2006,
author = {Bowler, Neill E. and Pierce, Clive E. and Seed, Alan W.},
title = {{STEPS}: A probabilistic precipitation forecasting scheme which merges an extrapolation nowcast with downscaled {NWP}},
journal = {Quarterly Journal of the Royal Meteorological Society},
volume = {132},
number = {620},
pages = {2127-2155},
doi = {https://doi.org/10.1256/qj.04.100},
url = {https://rmets.onlinelibrary.wiley.com/doi/abs/10.1256/qj.04.100},
eprint = {https://rmets.onlinelibrary.wiley.com/doi/pdf/10.1256/qj.04.100},
year = {2006}
}

@Article{Pulkkinen2019,
AUTHOR = {Pulkkinen, S. and Nerini, D. and P\'erez Hortal, A. A. and Velasco-Forero, C. and Seed, A. and Germann, U. and Foresti, L.},
TITLE = {Pysteps: an open-source Python library  for probabilistic precipitation nowcasting (v1.0)},
JOURNAL = {Geoscientific Model Development},
VOLUME = {12},
YEAR = {2019},
NUMBER = {10},
PAGES = {4185--4219},
URL = {https://gmd.copernicus.org/articles/12/4185/2019/},
DOI = {10.5194/gmd-12-4185-2019}
}

@article{Zhang2023,
  author = {Zhang, Yuchen and Long, Mingsheng and Chen, Kaiyuan and Xing, Lanxiang and Jin, Ronghua and Jordan, Michael I. and Wang, Jianmin},
  title = {Skilful nowcasting of extreme precipitation with {NowcastNet}},
  journal = {Nature},
  year = {2023},
  volume = {619},
  number = {7970},
  pages = {526--532},
  doi = {10.1038/s41586-023-06184-4},
  url = {https://doi.org/10.1038/s41586-023-06184-4}
}

@misc{Andrychowicz2023,
      title={Deep Learning for Day Forecasts from Sparse Observations}, 
      author={Marcin Andrychowicz and Lasse Espeholt and Di Li and Samier Merchant and Alexander Merose and Fred Zyda and Shreya Agrawal and Nal Kalchbrenner},
      year={2023},
      eprint={2306.06079},
      archivePrefix={arXiv},
      primaryClass={physics.ao-ph},
      url={https://arxiv.org/abs/2306.06079}, 
}

@article{Caseri2022,
title = {A convolutional recurrent neural network for strong convective rainfall nowcasting using weather radar data in Southeastern {B}razil},
journal = {Artificial Intelligence in Geosciences},
volume = {3},
pages = {8-13},
year = {2022},
issn = {2666-5441},
doi = {https://doi.org/10.1016/j.aiig.2022.06.001},
url = {https://www.sciencedirect.com/science/article/pii/S2666544122000211},
author = {Angelica N. Caseri and Leonardo Bacelar {Lima Santos} and Stephan Stephany}
}

@inproceedings{Shi2015,
 author = {Shi, Xingjian and Chen, Zhourong and Wang, Hao and Yeung, Dit-Yan and Wong, Wai-kin and WOO, Wang-chun},
 booktitle = {Advances in Neural Information Processing Systems},
 editor = {C. Cortes and N. Lawrence and D. Lee and M. Sugiyama and R. Garnett},
 pages = {},
 publisher = {Curran Associates, Inc.},
 title = {Convolutional LSTM Network: A Machine Learning Approach for Precipitation Nowcasting},
 url = {https://proceedings.neurips.cc/paper_files/paper/2015/file/07563a3fe3bbe7e3ba84431ad9d055af-Paper.pdf},
 volume = {28},
 year = {2015}
}

@article{Ravuri2021,
  author = {Ravuri, S. and Lenc, K. and Willson, M. and Kangin, D. and Lam, R. and Mirowski, P. and Fitzsimons, M. and Athanassiadou, M. and Kashem, S. and Madge, S. and Prudden, R. and Mandhane, A. and Clark, A. and Brock, A. and Simonyan, K. and Hadsell, R. and Robinson, N. and Clancy, E. and Arribas, A. and Mohamed, S.},
  title = {Skilful precipitation nowcasting using deep generative models of radar},
  journal = {Nature},
  volume = {597},
  number = {7878},
  pages = {672--677},
  year = {2021},
  doi = {10.1038/s41586-021-03854-z},
  url = {https://doi.org/10.1038/s41586-021-03854-z}
}

@misc{Goodfellow2014,
      title={Generative Adversarial Networks}, 
      author={Ian J. Goodfellow and Jean Pouget-Abadie and Mehdi Mirza and Bing Xu and David Warde-Farley and Sherjil Ozair and Aaron Courville and Yoshua Bengio},
      year={2014},
      eprint={1406.2661},
      archivePrefix={arXiv},
      primaryClass={stat.ML},
      url={https://arxiv.org/abs/1406.2661}, 
}

@article{Saxena2021,
author = {Saxena, Divya and Cao, Jiannong},
title = {Generative Adversarial Networks ({GANs}): Challenges, Solutions, and Future Directions},
year = {2021},
issue_date = {April 2022},
publisher = {Association for Computing Machinery},
address = {New York, NY, USA},
volume = {54},
number = {3},
issn = {0360-0300},
url = {https://doi.org/10.1145/3446374},
doi = {10.1145/3446374},
journal = {ACM Comput. Surv.},
month = may,
articleno = {63},
numpages = {42}
}

@InProceedings{Yu2024,
    author    = {Yu, Demin and Li, Xutao and Ye, Yunming and Zhang, Baoquan and Luo, Chuyao and Dai, Kuai and Wang, Rui and Chen, Xunlai},
    title     = {{DiffCast}: A Unified Framework via Residual Diffusion for Precipitation Nowcasting},
    booktitle = {Proceedings of the IEEE/CVF Conference on Computer Vision and Pattern Recognition (CVPR)},
    month     = {June},
    year      = {2024},
    pages     = {27758-27767}
}

@INPROCEEDINGS{Andrianakos2019,
  author={Andrianakos, George and Tsourounis, Dimitrios and Oikonomou, Spiros and Kastaniotis, Dimitris and Economou, George and Kazantzidis, Andreas},
  booktitle={2019 10th International Conference on Information, Intelligence, Systems and Applications (IISA)}, 
  title={Sky Image forecasting with Generative Adversarial Networks for cloud coverage prediction}, 
  year={2019},
  volume={},
  number={},
  pages={1-7},
  doi={10.1109/IISA.2019.8900774}}

@InProceedings{Rombach2022,
    author    = {Rombach, Robin and Blattmann, Andreas and Lorenz, Dominik and Esser, Patrick and Ommer, Bj\"orn},
    title     = {High-Resolution Image Synthesis With Latent Diffusion Models},
    booktitle = {Proceedings of the IEEE/CVF Conference on Computer Vision and Pattern Recognition (CVPR)},
    month     = {June},
    year      = {2022},
    pages     = {10684-10695}
}

@article{schmit2017,
  title={A closer look at the {ABI} on the {GOES-R} series},
  author={Schmit, Timothy J and Griffith, Paul and Gunshor, Mathew M and Daniels, Jaime M and Goodman, Steven J and Lebair, William J},
  journal={Bulletin of the American Meteorological Society},
  volume={98},
  number={4},
  pages={681--698},
  year={2017},
  publisher={American Meteorological Society}
}

@article{Zhang2018,
title = {Deep photovoltaic nowcasting},
journal = {Solar Energy},
volume = {176},
pages = {267-276},
year = {2018},
issn = {0038-092X},
doi = {https://doi.org/10.1016/j.solener.2018.10.024},
url = {https://www.sciencedirect.com/science/article/pii/S0038092X1831003X},
author = {Jinsong Zhang and Rodrigo Verschae and Shohei Nobuhara and Jean-François Lalonde}
}

@inproceedings{Craddock2024,
author = {Mary Ellen Craddock and Randall J. Alliss and Billy D. Felton and Heather L. Kiley},
title = {{The Lasercom atmospheric monitoring and prediction system}},
volume = {12877},
booktitle = {Free-Space Laser Communications XXXVI},
editor = {Hamid Hemmati and Bryan S. Robinson},
organization = {International Society for Optics and Photonics},
publisher = {SPIE},
pages = {1287716},
year = {2024},
doi = {10.1117/12.2692572},
URL = {https://doi.org/10.1117/12.2692572}
}

@Article{Kellerhals2022,
AUTHOR = {Kellerhals, Samuel A. and De Leeuw, Fons and Rodriguez Rivero, Cristian},
TITLE = {Cloud Nowcasting with Structure-Preserving Convolutional Gated Recurrent Units},
JOURNAL = {Atmosphere},
VOLUME = {13},
YEAR = {2022},
NUMBER = {10},
ARTICLE-NUMBER = {1632},
URL = {https://www.mdpi.com/2073-4433/13/10/1632},
ISSN = {2073-4433},
DOI = {10.3390/atmos13101632}
}

@article{Shakya2019,
author = {Shakya, Snehlata and Kumar, Sanjeev},
title = {Characterising and predicting the movement of clouds using fractional-order optical flow},
journal = {IET Image Processing},
volume = {13},
number = {8},
pages = {1375-1381},
doi = {https://doi.org/10.1049/iet-ipr.2018.6100},
url = {https://ietresearch.onlinelibrary.wiley.com/doi/abs/10.1049/iet-ipr.2018.6100},
eprint = {https://ietresearch.onlinelibrary.wiley.com/doi/pdf/10.1049/iet-ipr.2018.6100},
year = {2019}
}

@INPROCEEDINGS{Berthomier2020,
  author={Berthomier, Léa and Pradel, Bruno and Perez, Lior},
  booktitle={2020 Tenth International Conference on Image Processing Theory, Tools and Applications (IPTA)}, 
  title={Cloud Cover Nowcasting with Deep Learning}, 
  year={2020},
  volume={},
  number={},
  pages={1-6},
  doi={10.1109/IPTA50016.2020.9286606}}

@article {Griffin2017,
      author = "Sarah M. Griffin and Jason A. Otkin and Christopher M. Rozoff and Justin M. Sieglaff and Lee M. Cronce and Curtis R. Alexander and Tara L. Jensen and Jamie K. Wolff",
      title = "Seasonal Analysis of Cloud Objects in the High-Resolution Rapid Refresh ({HRRR}) Model Using Object-Based Verification",
      journal = "Journal of Applied Meteorology and Climatology",
      year = "2017",
      publisher = "American Meteorological Society",
      address = "Boston MA, USA",
      volume = "56",
      number = "8",
      doi = "10.1175/JAMC-D-17-0004.1",
      pages=      "2317 - 2334",
      url = "https://journals.ametsoc.org/view/journals/apme/56/8/jamc-d-17-0004.1.xml"
}

@article {Griffin2024,
      author = "Sarah M. Griffin and Jason A. Otkin and William E. Lewis",
      title = "Methods for Validating {HRRR} Simulated Cloud Properties for Different Weather Phenomena Using Satellite and Radar Observations",
      journal = "Weather and Forecasting",
      year = "2024",
      publisher = "American Meteorological Society",
      address = "Boston MA, USA",
      volume = "39",
      number = "1",
      doi = "10.1175/WAF-D-23-0109.1",
      pages=      "97 - 120",
      url = "https://journals.ametsoc.org/view/journals/wefo/39/1/WAF-D-23-0109.1.xml"
}

@article{Chen2023, title={{SwinRDM}: Integrate {SwinRNN} with Diffusion Model towards High-Resolution and High-Quality Weather Forecasting}, volume={37}, url={https://ojs.aaai.org/index.php/AAAI/article/view/25105}, DOI={10.1609/aaai.v37i1.25105}, abstractNote={Data-driven medium-range weather forecasting has attracted much attention in recent years. However, the forecasting accuracy at high resolution is unsatisfactory currently. Pursuing high-resolution and high-quality weather forecasting, we develop a data-driven model SwinRDM which integrates an improved version of SwinRNN with a diffusion model. SwinRDM performs predictions at 0.25-degree resolution and achieves superior forecasting accuracy to IFS (Integrated Forecast System), the state-of-the-art operational NWP model, on representative atmospheric variables including 500 hPa geopotential (Z500), 850 hPa temperature (T850), 2-m temperature (T2M), and total precipitation (TP), at lead times of up to 5 days. We propose to leverage a two-step strategy to achieve high-resolution predictions at 0.25-degree considering the trade-off between computation memory and forecasting accuracy. Recurrent predictions for future atmospheric fields are firstly performed at 1.40625-degree resolution, and then a diffusion-based super-resolution model is leveraged to recover the high spatial resolution and finer-scale atmospheric details. SwinRDM pushes forward the performance and potential of data-driven models for a large margin towards operational applications.}, number={1}, journal={Proceedings of the AAAI Conference on Artificial Intelligence}, author={Chen, Lei and Du, Fei and Hu, Yuan and Wang, Zhibin and Wang, Fan}, year={2023}, month={Jun.}, pages={322-330} }

@misc{Gao2023,
      title={{PreDiff}: Precipitation Nowcasting with Latent Diffusion Models}, 
      author={Zhihan Gao and Xingjian Shi and Boran Han and Hao Wang and Xiaoyong Jin and Danielle Maddix and Yi Zhu and Mu Li and Yuyang Wang},
      year={2023},
      eprint={2307.10422},
      archivePrefix={arXiv},
      primaryClass={cs.LG},
      url={https://arxiv.org/abs/2307.10422}, 
}

@article{Asperti2024,
  author = {Asperti, Andrea and Merizzi, Fabio and Paparella, Alberto and Pedrazzi, Giorgio and Angelinelli, Matteo and Colamonaco, Stefano},
  year = {2024},
  month = {December},
  title = {Precipitation Nowcasting with Generative Diffusion Models},
  journal = {Applied Intelligence},
  volume = {55},
  number = {3},
  pages = {187},
  doi = {10.1007/s10489-024-06048-y},
  url = {https://doi.org/10.1007/s10489-024-06048-y}
}

@article{Zhong2024,
  author = {Zhong, Xiaohui and Chen, Lei and Liu, Jun and Lin, Chensen and Qi, Yuan and Li, Hao},
  year = {2024},
  month = {December},
  title = {{FuXi-Extreme}: Improving Extreme Rainfall and Wind Forecasts with Diffusion Model},
  journal = {Science China Earth Sciences},
  volume = {67},
  number = {12},
  pages = {3696--3708},
  doi = {10.1007/s11430-023-1427-x},
  url = {https://doi.org/10.1007/s11430-023-1427-x}
}

@article{Li2024,
author = {Lizao Li  and Robert Carver  and Ignacio Lopez-Gomez  and Fei Sha  and John Anderson },
title = {Generative emulation of weather forecast ensembles with diffusion models},
journal = {Science Advances},
volume = {10},
number = {13},
pages = {eadk4489},
year = {2024},
doi = {10.1126/sciadv.adk4489},
URL = {https://www.science.org/doi/abs/10.1126/sciadv.adk4489},
eprint = {https://www.science.org/doi/pdf/10.1126/sciadv.adk4489}}

@inproceedings{Srivastava2024,
title={Precipitation Downscaling with Spatiotemporal Video Diffusion},
author={Prakhar Srivastava and Ruihan Yang and Gavin Kerrigan and Gideon Dresdner and Jeremy J McGibbon and Christopher S. Bretherton and Stephan Mandt},
booktitle={The Thirty-eighth Annual Conference on Neural Information Processing Systems},
year={2024},
url={https://openreview.net/forum?id=hhnkH8ex5d}
}

@InProceedings{Price2022,
  title = 	 { Increasing the accuracy and resolution of precipitation forecasts using deep generative models },
  author =       {Price, Ilan and Rasp, Stephan},
  booktitle = 	 {Proceedings of The 25th International Conference on Artificial Intelligence and Statistics},
  pages = 	 {10555--10571},
  year = 	 {2022},
  editor = 	 {Camps-Valls, Gustau and Ruiz, Francisco J. R. and Valera, Isabel},
  volume = 	 {151},
  series = 	 {Proceedings of Machine Learning Research},
  month = 	 {28--30 Mar},
  publisher =    {PMLR},
  url = 	 {https://proceedings.mlr.press/v151/price22a.html}
}

@article{Harris2022,
author = {Harris, Lucy and McRae, Andrew T. T. and Chantry, Matthew and Dueben, Peter D. and Palmer, Tim N.},
title = {A Generative Deep Learning Approach to Stochastic Downscaling of Precipitation Forecasts},
journal = {Journal of Advances in Modeling Earth Systems},
volume = {14},
number = {10},
pages = {e2022MS003120},
doi = {https://doi.org/10.1029/2022MS003120},
url = {https://agupubs.onlinelibrary.wiley.com/doi/abs/10.1029/2022MS003120},
eprint = {https://agupubs.onlinelibrary.wiley.com/doi/pdf/10.1029/2022MS003120},
note = {e2022MS003120 2022MS003120},
year = {2022}
}

@article{anderson1982reverse,
  title={Reverse-time diffusion equation models},
  author={Anderson, Brian DO},
  journal={Stochastic Processes and their Applications},
  volume={12},
  number={3},
  pages={313--326},
  year={1982},
  publisher={Elsevier}
}

@inproceedings{sohl2015deep,
  title={Deep unsupervised learning using nonequilibrium thermodynamics},
  author={Sohl-Dickstein, Jascha and Weiss, Eric and Maheswaranathan, Niru and Ganguli, Surya},
  booktitle={International conference on machine learning},
  pages={2256--2265},
  year={2015},
  organization={PMLR}
}

@misc{Ramesh2021,
      title={Zero-Shot Text-to-Image Generation}, 
      author={Aditya Ramesh and Mikhail Pavlov and Gabriel Goh and Scott Gray and Chelsea Voss and Alec Radford and Mark Chen and Ilya Sutskever},
      year={2021},
      eprint={2102.12092},
      archivePrefix={arXiv},
      primaryClass={cs.CV},
      url={https://arxiv.org/abs/2102.12092}, 
}

@misc{Saharia2022,
      title={Photorealistic Text-to-Image Diffusion Models with Deep Language Understanding}, 
      author={Chitwan Saharia and William Chan and Saurabh Saxena and Lala Li and Jay Whang and Emily Denton and Seyed Kamyar Seyed Ghasemipour and Burcu Karagol Ayan and S. Sara Mahdavi and Rapha Gontijo Lopes and Tim Salimans and Jonathan Ho and David J Fleet and Mohammad Norouzi},
      year={2022},
      eprint={2205.11487},
      archivePrefix={arXiv},
      primaryClass={cs.CV},
      url={https://arxiv.org/abs/2205.11487}, 
}

@misc{Ronneberger2015,
      title={{U-Net}: Convolutional Networks for Biomedical Image Segmentation}, 
      author={Olaf Ronneberger and Philipp Fischer and Thomas Brox},
      year={2015},
      eprint={1505.04597},
      archivePrefix={arXiv},
      primaryClass={cs.CV},
      url={https://arxiv.org/abs/1505.04597}, 
}

@misc{He2015,
      title={Deep Residual Learning for Image Recognition}, 
      author={Kaiming He and Xiangyu Zhang and Shaoqing Ren and Jian Sun},
      year={2015},
      eprint={1512.03385},
      archivePrefix={arXiv},
      primaryClass={cs.CV},
      url={https://arxiv.org/abs/1512.03385}, 
}

@article {Haynes2023,
      author = "Katherine Haynes and Ryan Lagerquist and Marie McGraw and Kate Musgrave and Imme Ebert-Uphoff",
      title = "Creating and Evaluating Uncertainty Estimates with Neural Networks for Environmental-Science Applications",
      journal = "Artificial Intelligence for the Earth Systems",
      year = "2023",
      publisher = "American Meteorological Society",
      address = "Boston MA, USA",
      volume = "2",
      number = "2",
      doi = "10.1175/AIES-D-22-0061.1",
      pages=      "220061",
      url = "https://journals.ametsoc.org/view/journals/aies/2/2/AIES-D-22-0061.1.xml"
}

@misc{Xu2025,
      title={One-step Diffusion Models with $f$-Divergence Distribution Matching}, 
      author={Yilun Xu and Weili Nie and Arash Vahdat},
      year={2025},
      eprint={2502.15681},
      archivePrefix={arXiv},
      primaryClass={cs.LG},
      url={https://arxiv.org/abs/2502.15681}, 
}

@article{Wang2024,
author = {Wang, Rui and Fung, Jimmy C. H. and Lau, Alexis K. H.},
title = {Skillful Precipitation Nowcasting Using Physical-Driven Diffusion Networks},
journal = {Geophysical Research Letters},
volume = {51},
number = {24},
pages = {e2024GL110832},
doi = {https://doi.org/10.1029/2024GL110832},
url = {https://agupubs.onlinelibrary.wiley.com/doi/abs/10.1029/2024GL110832},
eprint = {https://agupubs.onlinelibrary.wiley.com/doi/pdf/10.1029/2024GL110832},
note = {e2024GL110832 2024GL110832},
year = {2024}
}

@misc{Dai2024,
      title={Four-hour thunderstorm nowcasting using deep diffusion models of satellite}, 
      author={Kuai Dai and Xutao Li and Junying Fang and Yunming Ye and Demin Yu and Di Xian and Danyu Qin and Jingsong Wang},
      year={2024},
      eprint={2404.10512},
      archivePrefix={arXiv},
      primaryClass={cs.LG},
      url={https://arxiv.org/abs/2404.10512}, 
}

@article{Nai2024,
doi = {10.1088/1748-9326/ad2891},
url = {https://dx.doi.org/10.1088/1748-9326/ad2891},
year = {2024},
month = {feb},
publisher = {IOP Publishing},
volume = {19},
number = {3},
pages = {034039},
author = {Nai, Congyi and Pan, Baoxiang and Chen, Xi and Tang, Qiuhong and Ni, Guangheng and Duan, Qingyun and Lu, Bo and Xiao, Ziniu and Liu, Xingcai},
title = {Reliable precipitation nowcasting using probabilistic diffusion models},
journal = {Environmental Research Letters}
}

@misc{Hatanaka2023,
      title={Diffusion Models for High-Resolution Solar Forecasts}, 
      author={Yusuke Hatanaka and Yannik Glaser and Geoff Galgon and Giuseppe Torri and Peter Sadowski},
      year={2023},
      eprint={2302.00170},
      archivePrefix={arXiv},
      primaryClass={cs.LG},
      url={https://arxiv.org/abs/2302.00170}, 
}

@article { dellemonache2013probabilistic,
      author = "Delle Monache, Luca and F. Anthony Eckel and Daran L. Rife and Badrinath Nagarajan and Keith Searight",
      title = "Probabilistic Weather Prediction with an Analog Ensemble",
      journal = "Monthly Weather Review",
      year = "2013",
      publisher = "American Meteorological Society",
      address = "Boston MA, USA",
      volume = "141",
      number = "10",
      doi = "10.1175/MWR-D-12-00281.1",
      pages=      "3498 - 3516",
      url = "https://journals.ametsoc.org/view/journals/mwre/141/10/mwr-d-12-00281.1.xml"
}

@article{sharpness2025,
      author = {Ebert-Uphoff, Imme and  Ver Hoef, Lander and Schreckk, John S. and Stock, Jason and Molina, Maria J. and McGovern, Amy and Yu, Michael and Petzke, Bill and Hilburn, Kyle and Hall, David M. and Gagne, David J. and Campbell, William F. and Radford, Jacob T. and Stewart, Jebb Q. and Scheuerman, Sam},
      title = "Measuring Sharpness of {AI}-Generated Meteorological Imagery",
      journal = "Artificial Intelligence for the Earth Systems (in press)",
      year = "2025",
      publisher = "American Meteorological Society"
}

@misc{Martin2025,
  title        = {Generative Data Assimilation for Surface Ocean State Estimation from Multi-Modal Satellite Observations},
  author       = {Martin, Scott and Manucharyan, Georgy and Klein, Patrice},
  year         = {2025},
  month        = {March},
  publisher    = {EarthArXiv},
  doi          = {https://doi.org/10.31223/X5ZT6N},
  url          = {https://eartharxiv.org/repository/view/8688/},
  note         = {Preprint}
}

@InProceedings{Qu_2024,
    author    = {Qu, Yongquan and Nathaniel, Juan and Li, Shuolin and Gentine, Pierre},
    title     = {Deep Generative Data Assimilation in Multimodal Setting},
    booktitle = {Proceedings of the IEEE/CVF Conference on Computer Vision and Pattern Recognition (CVPR) Workshops},
    month     = {June},
    year      = {2024},
    pages     = {449-459}
}

@misc{heidinger2018enterprise,
  title={Enterprise {AWG} {Cloud} {Height} {Algorithm} ({ACHA}) - Version 3.4},
  author={Heidinger, AK and Li, Y and Wanzong, S},
  note={NOAA NESDIS Center for Satellite Applications and Research: Washington, DC, USA},
  howpublished = {Available at \url{https://www.star.nesdis.noaa.gov/jpss/documents/ATBD/ATBD_EPS_Cloud_ACHA_v3.4.pdf}},
  month = {Sept},
  year={2020}
}

@article{bouallegue2024rise,
  title={The rise of data-driven weather forecasting: A first statistical assessment of machine learning--based weather forecasts in an operational-like context},
  author={Bouall{\`e}gue, Zied Ben and Clare, Mariana CA and Magnusson, Linus and Gascon, Estibaliz and Maier-Gerber, Michael and Janou{\v{s}}ek, Martin and Rodwell, Mark and Pinault, Florian and Dramsch, Jesper S and Lang, Simon TK and others},
  journal={Bulletin of the American Meteorological Society},
  volume={105},
  number={6},
  pages={E864--E883},
  year={2024},
  publisher={American Meteorological Society}
}

@misc{Andry2025,
      title={Appa: Bending Weather Dynamics with Latent Diffusion Models for Global Data Assimilation}, 
      author={Gérôme Andry and François Rozet and Sacha Lewin and Omer Rochman and Victor Mangeleer and Matthias Pirlet and Elise Faulx and Marilaure Grégoire and Gilles Louppe},
      year={2025},
      eprint={2504.18720},
      archivePrefix={arXiv},
      primaryClass={cs.LG},
      url={https://arxiv.org/abs/2504.18720}, 
}

@misc{Couairon2024,
      title={ArchesWeather \& ArchesWeatherGen: a deterministic and generative model for efficient ML weather forecasting}, 
      author={Guillaume Couairon and Renu Singh and Anastase Charantonis and Christian Lessig and Claire Monteleoni},
      year={2024},
      eprint={2412.12971},
      archivePrefix={arXiv},
      primaryClass={cs.LG},
      url={https://arxiv.org/abs/2412.12971}, 
}

@misc{Lang2024a,
      title={AIFS -- ECMWF's data-driven forecasting system}, 
      author={Simon Lang and Mihai Alexe and Matthew Chantry and Jesper Dramsch and Florian Pinault and Baudouin Raoult and Mariana C. A. Clare and Christian Lessig and Michael Maier-Gerber and Linus Magnusson and Zied Ben Bouallègue and Ana Prieto Nemesio and Peter D. Dueben and Andrew Brown and Florian Pappenberger and Florence Rabier},
      year={2024},
      eprint={2406.01465},
      archivePrefix={arXiv},
      primaryClass={physics.ao-ph},
      url={https://arxiv.org/abs/2406.01465}, 
}

@misc{Lang2024b,
      title={AIFS-CRPS: Ensemble forecasting using a model trained with a loss function based on the Continuous Ranked Probability Score}, 
      author={Simon Lang and Mihai Alexe and Mariana C. A. Clare and Christopher Roberts and Rilwan Adewoyin and Zied Ben Bouallègue and Matthew Chantry and Jesper Dramsch and Peter D. Dueben and Sara Hahner and Pedro Maciel and Ana Prieto-Nemesio and Cathal O'Brien and Florian Pinault and Jan Polster and Baudouin Raoult and Steffen Tietsche and Martin Leutbecher},
      year={2024},
      eprint={2412.15832},
      archivePrefix={arXiv},
      primaryClass={physics.ao-ph},
      url={https://arxiv.org/abs/2412.15832}, 
}

@inproceedings{Chantry2025AIFS,
  author       = {Chantry, Matthew and Lang, Simon and Alexe, Mihai and Dramsch, Jesper and Raoult, Baudouin and Clare, Mariana and Santa Cruz, Mario and Hahner, Sara and Adewoyin, Rilwan and Pinault, Florian and Prieto Nemesio, Ana and Moldovan, Gabriel and Magnusson, Linus and Ben Bouallegue, Zied and Tietsche, Steffen and Pinnington, Ewan Mark and Schloer, Jakob and Brown, Andy and Pappenberger, Florian and Rabier, Florence},
  title        = {AIFS -- ECMWF’s Data-Driven Forecasting System},
  booktitle    = {Proceedings of the 105th Annual Meeting of the American Meteorological Society},
  year         = {2025},
  address      = {New Orleans, LA},
  month        = jan,
  pages        = {–},
  note         = {Paper ID 449087; Meeting held 12–16 Jan 2025}
}

@misc{Alet2025,
      title={Skillful joint probabilistic weather forecasting from marginals}, 
      author={Ferran Alet and Ilan Price and Andrew El-Kadi and Dominic Masters and Stratis Markou and Tom R. Andersson and Jacklynn Stott and Remi Lam and Matthew Willson and Alvaro Sanchez-Gonzalez and Peter Battaglia},
      year={2025},
      eprint={2506.10772},
      archivePrefix={arXiv},
      primaryClass={cs.LG},
      url={https://arxiv.org/abs/2506.10772}, 
}

@article{Zhong2025,
author = {Xiaohui Zhong  and Lei Chen  and Hao Li  and Roberto Buizza  and Jun Liu  and Jie Feng  and Zijian Zhu  and Xu Fan  and Kan Dai  and Jing-jia Luo  and Jie Wu  and Bo Lu },
title = {FuXi-ENS: A machine learning model for efficient and accurate ensemble weather prediction},
journal = {Science Advances},
volume = {11},
number = {44},
pages = {eadu2854},
year = {2025},
doi = {10.1126/sciadv.adu2854},
URL = {https://www.science.org/doi/abs/10.1126/sciadv.adu2854},
eprint = {https://www.science.org/doi/pdf/10.1126/sciadv.adu2854}}

@misc{Lipman2023,
      title={Flow Matching for Generative Modeling}, 
      author={Yaron Lipman and Ricky T. Q. Chen and Heli Ben-Hamu and Maximilian Nickel and Matt Le},
      year={2023},
      eprint={2210.02747},
      archivePrefix={arXiv},
      primaryClass={cs.LG},
      url={https://arxiv.org/abs/2210.02747}, 
}

@misc{Tong2024,
      title={Improving and generalizing flow-based generative models with minibatch optimal transport}, 
      author={Alexander Tong and Kilian Fatras and Nikolay Malkin and Guillaume Huguet and Yanlei Zhang and Jarrid Rector-Brooks and Guy Wolf and Yoshua Bengio},
      year={2024},
      eprint={2302.00482},
      archivePrefix={arXiv},
      primaryClass={cs.LG},
      url={https://arxiv.org/abs/2302.00482}, 
}

@misc{Stock2025,
      title={Swift: An Autoregressive Consistency Model for Efficient Weather Forecasting}, 
      author={Jason Stock and Troy Arcomano and Rao Kotamarthi},
      year={2025},
      eprint={2509.25631},
      archivePrefix={arXiv},
      primaryClass={cs.LG},
      url={https://arxiv.org/abs/2509.25631}, 
}

@InProceedings{Isola2017,
    author    = {Isola, P and Zhu, J.-Y. and Efros, A. A.},
    title     = {Image-to-Image Translation with Conditional Adversarial Networks},
    booktitle = {2017 {IEEE} Conference on Computer Vision and Pattern Recognition (CVPR)},
    month     = {},
    year      = {2017},
    pages     = {5967-5976}
}

\end{document}